\documentclass{article}

\usepackage[preprint]{neurips_2026}

\usepackage[utf8]{inputenc} 
\usepackage[T1]{fontenc}    
\usepackage{hyperref}       
\usepackage{url}            
\usepackage{booktabs}       
\usepackage{amsfonts}       
\usepackage{nicefrac}       
\usepackage{microtype}      
\usepackage{xcolor}         
\usepackage{amsthm}
\usepackage{amsmath}
\usepackage{natbib}
\usepackage{comment}       
\usepackage{graphicx}
\usepackage{amssymb}
\usepackage{tabularx}
\usepackage{tabularray}
\UseTblrLibrary{booktabs}

\theoremstyle{plain}

\theoremstyle{definition}

\theoremstyle{remark}
\newtheorem{remark}{Remark}

\usepackage{algorithm}
\usepackage{algpseudocode}
\usepackage{mathtools}                 

\newcommand{\Pp}{\mathbb{P}}
\newcommand{\E}{\mathbb{E}}
\newcommand{\R}{\mathbb{R}}
\newcommand{\X}{\mathcal{X}}
\newcommand{\D}{\mathcal{D}}
\DeclareMathOperator{\conv}{conv}
\DeclareMathOperator{\rank}{rank}
\DeclareMathOperator{\supp}{supp}
\DeclareMathOperator{\diag}{diag}

\newcounter{inlineassump}
\renewcommand{\theinlineassump}{A\arabic{inlineassump}}

\newcounter{inlinethm}
\renewcommand{\theinlinethm}{\arabic{inlinethm}}

\newcommand{\asmp}[1]{\ref{#1}} 
\newcommand{\setasmp}[2]{%
    \refstepcounter{inlineassump}
    \label{#1}
    \textbf{(\theinlineassump)} \emph{#2}
}

\newcommand{\setthm}[2]{%
    \refstepcounter{inlinethm}
    \label{#1}
    \textbf{Theorem~\theinlinethm\ (#2).}
}

\title{Multiclass Classification without Labels via Posterior Simplex Geometry}
\author{%
  Rapha{\"e}l Bonnet-Guerrini\thanks{Primary authors - ordered alphabetically, contributed equally. Led the project, developed the methodology, implemented the experiments, analyzed the results, and wrote the manuscript.} \\
  Computer Science Department\\
  Universit\`a degli Studi di Milano\\
  Istituto Nazionale di Fisica Nucleare\\ 
  20133 Milano, Italy \\
  \texttt{raphael.bonnet-guerrini@unimi.it} \\
  \And
  Johann Ioannou-Nikolaides\footnotemark[1] \\
  Niels Bohr Institute \\
  University of Copenhagen \\
  Copenhagen, 2100 \\
  \texttt{johann.nikolaides@nbi.ku.dk}
  \And
  Troels Petersen\thanks{Supervising authors.} \\
  Niels Bohr Institute \\
  University of Copenhagen \\
  Copenhagen, 2100 \\ 
  \And
  Vincenzo Piuri\footnotemark[2] \\
  Computer Science Department\\
  Universit\`a degli Studi di Milano\\
  Milan, MI, 20133 \\
}

\begin{document}

\maketitle

\begin{abstract}In many classification problems, reliable instance-level labels are unavailable. However, it is often possible to construct weakly enriched unlabeled samples: datasets selected by different cuts, sources, populations, or experimental conditions that change latent class proportions without revealing them. Classification without Labels (CWoLa) shows that, in the binary case ($K=2$), a classifier trained to distinguish two impure mixtures with different class proportions can recover an optimal class discriminator without knowing the mixture proportions. We extend this principle to multiclass learning from several unlabeled mixtures ($K>2$), where the learner observes only mixture identity and neither latent class labels nor class-prior matrices. We prove that, for a multiclass mixture model, the Bayes-optimal mixture classifier $g^\star$ maps data points into a $(K-1)$-simplex embedded in mixture-posterior space. The $K$ vertices of this simplex are induced by the latent classes through the unknown mixing matrix. Leveraging this geometry, we propose prior-free procedures that train a standard classifier to distinguish mixture identities and then extract latent class structure using either post-hoc simplex fitting or a bottleneck architecture. Experiments on MNIST, CIFAR-10, and Galaxy10 DECaLS show that mixture identity alone can recover latent classes and their fractions in the mixture. By narrowing the gap between weakly supervised and fully supervised performance, we provide a mathematically grounded, scalable tool for multiclass discovery in label-scarce domains.
\end{abstract}


\section{Introduction}

Deep neural networks have achieved strong empirical performance across a wide range of domains~\citep{lecun:hal-04206682}. 
In classification, however, these gains often rely on large datasets with reliable instance-level labels, a requirement that remains a major bottleneck in practice~\citep{Russakovsky2014ImageNetLS,Ratner2017SnorkelRT}. 
This limitation is especially acute in scientific and clinical domains, where labels may require large-scale expert annotation or human inspection~\citep{Darg:2009rc,Wang2020AnnotationefficientDL}
(e.g., community annotations from Zooniverse are scalable but weak).

Several paradigms reduce the dependence on clean instance-level labels, but under different assumptions. 
Self-supervised methods learn useful representations from unlabeled data, yet typically require task-specific labels to define semantic classes~\citep{Chen2020ASF,He2021MaskedAA}. 
Robust-learning methods tolerate corrupted labels, but still assume instance-level labels are observed during training~\citep{Han2018CoteachingRT}. 
Closest to our setting, multiclass learning from multiple unlabeled datasets exploits variation in class proportions across mixtures, but typically assumes that these proportions are known~\citep{Tang2022MulticlassCF,wei2024consistent}.

\begin{figure}
    \centering
    \includegraphics[width=1\linewidth]{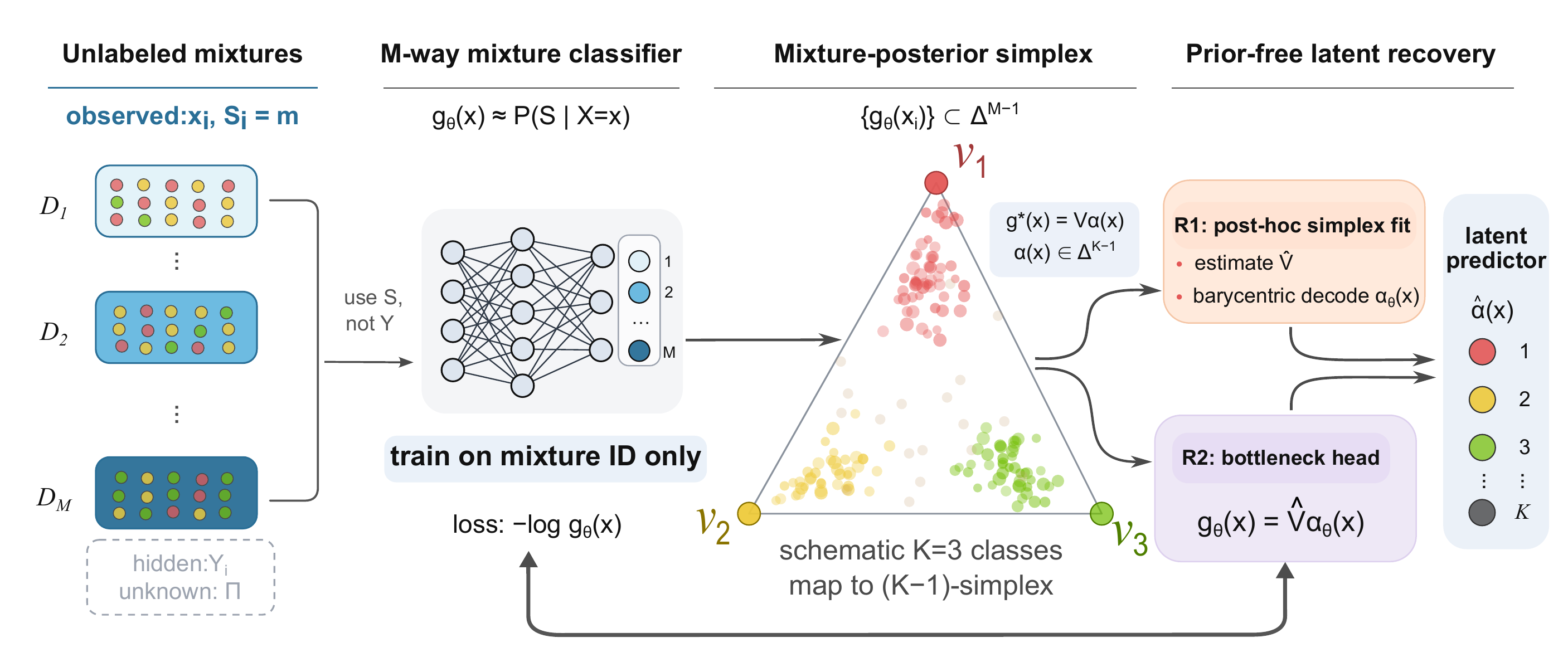}
    \caption{
    Overview of posterior simplex geometry via multiclass CWoLa.
    From unlabeled mixtures with observed mixture identities only, we train an \(M\)-way classifier and use its source-posterior geometry to recover latent class structure.
    The learned posterior cloud lies in a \((K-1)\)-simplex whose vertices correspond to the latent classes.
    Prior-free recovery is performed either by post-hoc simplex fitting or by an architectural bottleneck.
    Colors indicate latent classes for schematic illustration only and are not used during training.
    }
    \label{fig:global-schematic}
\end{figure}
We study a setting with \(K\) latent classes and \(M\) observed mixtures. 
Each example has an unobserved class \(y \in \{1,\dots,K\}\) and an observed mixture identity \(m \in \{1,\dots,M\}\). 
Mixture \(m\) has density
\(
q_m(x)=\sum_{k=1}^K \pi_{mk} p_k(x),
\)
where the class-conditionals \(p_k\) are shared across mixtures and the mixing weights \(\pi_{mk}\) are unknown. 
The learner observes \(x\) and \(m\), but not \(y\) or the mixing matrix \(\Pi=(\pi_{mk})\). 
The mixtures may arise naturally across data-collection contexts, or be deliberately induced by weak cuts that enrich or deplete latent classes.
Importantly, they do not need to be clean, calibrated, nor high-purity. In many applications, the key quantity is not only an instance-level label but the latent composition of each mixture itself. Recovering the posterior simplex identifies the hidden mixing matrix, enabling estimation of mixture-level class proportions without observing them.

This setting is inspired by Classification without Labels (CWoLa). 
In the binary case, CWoLa shows that, given two mixtures of the same two latent classes with different class proportions, the optimal classifier trained to distinguish the mixtures is also optimal for distinguishing the underlying classes, up to the orientation of the score~\citep{Metodiev:2017vrx}. 
Thus, CWoLa turns mixture discrimination into class discrimination without requiring instance-level labels or mixture proportions. 
However, the original guarantee is binary and does not directly provide a multiclass extension when several mixtures contain different proportions of more than two latent classes.

In this work, we develop a prior-free framework for multiclass CWoLa. 
Our starting point is simple: train a classifier to predict the mixture identity of each example. 
Our main contributions are:
\begin{itemize}
    \item We show that the Bayes-optimal mixture posterior $g^{\star}(x) = \mathbb{P}(m \mid x)$ lies in a $(K-1)$-simplex inside $\Delta^{M-1}$, whose vertices are induced by the latent classes through the unknown rectangular mixing matrix.
    \item We prove that this posterior geometry is sufficient to recover both the latent class posterior and the mixture composition matrix, up to permutation and prior reweighting, without observing instance-level labels or mixture proportions.    
    \item We propose prior-free recovery procedures based on post-hoc simplex fitting and an architectural bottleneck, and evaluate them on MNIST, Fashion-MNIST, CIFAR-10, and Galaxy10 DECaLS.
\end{itemize}

To facilitate the application of our methods beyond the experiments considered here, we release \texttt{MultiCWoLa}, a reusable library implementing both the post-hoc and bottleneck approaches and designed to simplify their integration into other application-specific learning pipelines.\footnote{\href{https://github.com/rbonnetguerrini/MultiCWoLa}{\texttt{MultiCWoLa repository}}}

\paragraph{Note added}
While finalizing our analysis, we learned of \citep{deLaFuente:SimplexDemixing}, which also proposes a simplex method for demixing multiple unlabeled samples and applies it to light-flavor jet identification at colliders. The overall mathematical framework is consistent with ours, though we use complementary strategies for identifying the simplex geometry: simplex regularization via loss terms in \citep{deLaFuente:SimplexDemixing} versus post-hoc simplex fitting and architectural bottlenecks in Sec.~\ref{sec:methodology} below.

\section{Related Work}

\textbf{Classification without Labels.}
Our work is most directly inspired by Classification without Labels (CWoLa)~\citep{Metodiev:2017vrx}, which shows in the binary case that mixture discrimination can recover the optimal class discriminator without clean labels or known mixture proportions.
CWoLa has been used in high-energy physics for classification from impure samples, including quark--gluon discrimination and resonance/anomaly searches~\citep{Komiske2018LearningTC,Collins2018CWoLaHE}.
Existing guarantees, however, are fundamentally binary: they yield a one-dimensional class ordering rather than a multiclass representation.
Rather than reducing the problem to one-vs-rest (OvR) classifiers~\citep{Rifkin2004InDO}, we study the full \(M\)-way mixture posterior and show that it contains a latent simplex geometry.

\textbf{Learning from label proportions and multiple unlabeled datasets.}
Learning from label proportions (LLP) and multiclass classification from multiple unlabeled datasets also exploit variation in class proportions across bags or sources~\citep{Scott2020LearningFL,Tang2022MulticlassCF,wei2024consistent}.
These methods typically assume that each bag is accompanied by its class-proportion vector, or that the class-prior matrix is known, enabling unbiased or consistent risk estimators.
Our setting removes this supervision: the learner observes only mixture identity, with neither instance-level labels nor mixture proportions.
We therefore include known-prior methods as informative references, but not as direct fair competitors.

\textbf{Simplex geometry, separable NMF, and topic models.}
The simplex structure we identify is related to separable nonnegative matrix factorization and topic modeling, where latent components become identifiable as extreme rays, anchors, or vertices of a simplicial structure~\citep{cutler1994archetypal,Donoho2003:nmf,Arora2011ComputingAN,Arora2012LearningTM,Arora2012APA,Ke2017UsingSF}.
Our separability assumption plays an analogous role in mixture-posterior space.
The contribution is not a new vertex-identifiability principle in isolation, but the identification of the relevant discriminative object for multiclass CWoLa: the Bayes-optimal mixture posterior \(g^\star(x)=P(m\mid x)\).
Unlike topic models, where the simplex is built from word-document statistics, our simplex emerges from a classifier trained only to predict observable mixture identity (See Fig.~\ref{fig:global-schematic}).
Anchor-free topic models and minimum-volume methods in hyperspectral unmixing show that simplex recovery may be possible beyond pure-anchor or pure-pixel assumptions~\citep{Chan2009ASF,BioucasDias2012HyperspectralUO,Huang2016AnchorFreeCT}; our post-hoc fitting methods - in particular the constrained-$\hat\Pi$ and spread-initialised archetypal variants - are designed with this in mind.

\textbf{Decontamination and mutual contamination.} Our setting is also related to mutual contamination, mixture-proportion estimation, and decontamination, which study when base distributions or contamination proportions can be recovered from contaminated samples~\citep{KatzSamuels2017DecontaminationOM,Scott2020LearningFL}.  

\textbf{Learning with class-conditional label noise.}
In label-noise learning, a noisy label is generated from the clean label through an unknown transition matrix~\citep{Scott2013ClassificationWA,Liu2015}.
Recent methods estimate the transition matrix directly from noisy data, via total-variation regularization~\citep{Zhang2021LearningNT} or by minimizing the volume of the simplex enclosing the noisy posteriors~\citep{Li2021ProvablyEL}.
When $M=K$, our setting is closely related to class-conditional label noise if mixture identity is interpreted as a noisy label. In both cases, the observable-label posterior has a latent simplex structure, but our formulation additionally allows rectangular, overcomplete mixtures ($M>K$) with no label semantics.
Unlike methods that rely on a dominant clean-label component, our mixtures need not contain a dominant class: latent classes are identified through variation in their proportions across mixtures, connecting the setting to enrichment-based anomaly detection~\citep{Metodiev:2017vrx,Collins2018CWoLaHE}.

\section{Multiclass CWoLa}
\label{sec:methodology}

We develop the theoretical core of multiclass CWoLa by addressing two questions: (Q1) \emph{What is the geometry of the optimal mixture-identity classifier?} and (Q2) \emph{How do we extract this geometry from data?} We show that the optimal source posterior is constrained to a $(K-1)$-simplex inside $\Delta^{M-1}$, allowing us to recover latent posteriors via post-hoc vertex hunting or end-to-end bottlenecks.

\paragraph{Setup and Assumptions.} 
We observe $M$ unlabeled mixtures $\D_m$ composed of $K$ latent classes with densities $p_1, \dots, p_K$. Following the mixture model, the observed density of mixture $m$ is:
\begin{equation}
\label{eq:mixture-model}
q_m(x) \;=\; \sum_{k=1}^K \pi_{mk}\,p_k(x)
\end{equation}
where $\Pi = (\pi_{mk}) \in \R^{M\times K}$ is a row-stochastic mixing matrix. Let $S \in \{1, \dots, M\}$ denote the mixture identity and $Y \in \{1, \dots, K\}$ the latent class. We define the \emph{Bayes-optimal mixture posterior} $g^\star: \X \to \Delta^{M-1}$ as $g_m^\star(x) := \Pp(S=m \mid X=x)$ and the \emph{latent posterior} $h^\star: \X \to \Delta^{K-1}$ as $h_k^\star(x) := \Pp(Y=k \mid X=x)$. Our goal is to recover $h^\star$ (up to a permutation) from samples $\{(x_i, m_i)\}_{i=1}^n$. 

Our analysis relies on four conditions: 
\setasmp{ass:shared}{Shared class-conditionals}: $p_k$ is independent of $m$; 
\setasmp{ass:rank}{Full column rank}: $\rank(\Pi)=K$, implying $M \ge K$; 
\setasmp{ass:irreducibility}{Separability}: each class $k$ has an anchor region $\X_k$ carrying positive class-conditional mass, $\Pp(X\in\X_k \mid Y=k)>0$, on which $p_k(x) > 0$ and $p_{j \ne k}(x) = 0$; 
\setasmp{ass:uniform}{Uniform sampling}: $\Pp(S=m)=1/M$.
These conditions are standard in topic modeling \citep{Arora2012LearningTM} and mixture-label theory \citep{KatzSamuels2017DecontaminationOM}. A1 makes the problem well posed. A2 ensures the simplex covers $K-1$ dimensions, while A3 guarantees the data points reach its vertices. A4 is a normalization convenience rather than a substantive restriction: if the mixtures have unequal sizes, every statement below holds with the observed sampling frequencies $\Pp(S=m)$ in place of $1/M$, at the cost of slightly heavier notation.

\subsection{Posterior simplex geometry}
\label{sec:simplex-geometry}

This section addresses Q1 by characterizing the relationship between the observed mixture posterior $g^\star$ and the latent class posterior $h^\star$. We show that the image of $g^\star$ is geometrically constrained to a simplex whose vertices reveal the hidden mixing matrix $\Pi$. The proofs can be found in Appendices~\ref{app:proofs:simplex}, \ref{app:proofs:sufficient}, \ref{app:proofs:identifiability}.

\setthm{thm:simplex}{Posterior Simplex Theorem}
\emph{Under \asmp{ass:shared}, \asmp{ass:rank}, and \asmp{ass:uniform}, define the class abundances $c_k := \sum_{m=1}^M \pi_{mk}$, the class vertices $v_k := \frac{1}{c_k}\Pi_{:,k} \in \Delta^{M-1}$, and the barycentric weights}
\begin{equation*}
\alpha_k(x) \;:=\; \frac{c_k\,p_k(x)}{\sum_{\ell=1}^K c_\ell\,p_\ell(x)},
\qquad\text{which satisfy } \alpha_k(x)\in[0,1],\ \ \sum_{k=1}^K \alpha_k(x)=1.
\end{equation*}
\emph{Then, for $\mu_X$-almost every $x\in\X$,}
\begin{equation}
\label{eq:simplex-factorization}
g^\star(x) \;=\; \sum_{k=1}^K \alpha_k(x)\,v_k.
\end{equation}
In matrix form, $g^\star(x) = V\alpha(x)$ where $V=[v_1,\dots,v_K] \in \R^{M\times K}$. The vertices $v_k$ are affinely independent, so their convex hull $\mathcal{S} := \conv\{v_1,\dots,v_K\}$ is a $(K-1)$-simplex in $\Delta^{M-1}$ containing the image of $g^\star$. Moreover $\alpha(x)$ is exactly the latent class posterior of the pooled training population (Theorem~\ref{thm:sufficient-statistic}).

\setthm{thm:sufficient-statistic}{Multiclass CWoLa Optimality}
\emph{Under \asmp{ass:shared}, \asmp{ass:rank}, and \asmp{ass:uniform}, let $\bar h^\star(x) := (\alpha_1(x),\dots,\alpha_K(x))^\top$. Then $\bar h^\star$ is the latent posterior under the prior $\bar\pi_k := c_k/M$, and $g^\star(x) = V \bar h^\star(x)$. Because $V$ has full column rank, $\bar h^\star$ is recoverable via the linear map:}
\begin{equation}
\label{eq:inverse-map}
\bar h^\star(x) \;=\; (V^\top V)^{-1} V^\top g^\star(x).
\end{equation}
We call $\bar\pi$ the \emph{effective prior} as it is not an extra modeling choice but simply the class prior of the pooled training population, $\bar\pi_k=\Pp(Y=k)$ when the $M$ mixtures are pooled with equal weights, so $\bar h^\star_k(x)=\Pp(Y=k\mid X=x)$ on the data one actually trains on. The posterior $h^\star$ under any other class prior (e.g., a deployment population with different class frequencies) is a coordinate-wise reweighting of $\bar h^\star$. In particular $h^\star$ is a deterministic function of $g^\star$, so $Y \perp X \mid g^\star(X)$: $g^\star(x)$ is predictively (Bayes-) sufficient for the latent class $Y$, and the optimal latent-class classifier can be recovered from the optimal mixture classifier $g^\star(x)$.

\setthm{thm:identifiability}{Vertex Identifiability}
\emph{Let $g^\star_\#\mu_X$ denote the distribution of $g^\star(X)$ when $X\sim\mu_X$ (the pooled data distribution), and $\supp(g^\star_\#\mu_X)$ its support, i.e.\ the smallest closed set carrying the full probability mass of the posterior cloud. Under \asmp{ass:shared}-\asmp{ass:uniform}: (i) for $\mu_X$-almost every anchor point $x \in \X_k$,
\begin{equation}
    g^\star(x) = v_k;
\end{equation}
(ii) the vertices $\{v_k\}_{k=1}^K$ are exactly the extreme points of the convex hull of $\supp(g^\star_\#\mu_X)$, whose hull is the smallest closed simplex containing $\supp(g^\star_\#\mu_X)$.} Consequently, $V$ is identifiable up to a permutation of its columns from the population law of $g^\star(X)$, and $\Pi$ follows: the abundances are the unique solution of the linear system $Vc=\mathbf{1}_M$, and $\Pi = V \diag(c)$.\footnote{The identifiability statement is the posterior-space analogue of the residue-operator result in \citep{KatzSamuels2017DecontaminationOM}.}

\textbf{Geometric Intuition.}
The image $g^\star(\X)$ is confined to a $(K-1)$-dimensional simplex $\mathcal{S}$ embedded in the larger $(M-1)$-dimensional source-posterior space. The simplex representation contains the same information as the optimal latent-class classifier. When \asmp{ass:irreducibility} (Separability) holds, the data points in the anchor regions lie directly on the corners of this simplex. This transforms the problem of learning labels into a problem of vertex hunting. After finding the corners $v_k$ of the cloud of points $g^\star(x_i)$, Eq.~\ref{eq:inverse-map} provides the labels. The same vertices also determine the hidden mixing matrix via rescaling of the vertex points with the class abundances, and therefore recover the latent class proportions within each mixture. \newline
In summary, the geometric perspective simultaneously enables recovery of the latent class posteriors and the hidden class composition of each mixture.

\subsection{Practical Recovery and Limitations}
\label{sec:recovery-limitations}

In practice we estimate $g^\star$ by training an $M$-way classifier $g_\theta$ via the cross-entropy loss. Once a $g_\theta$ is fit, two complementary procedures convert $g_\theta$ into a $K$-class predictor:\newline
\textbf{(R1) Post-hoc simplex fitting.} Train an unconstrained $M$-way classifier $g_\theta:\X\to\Delta^{M-1}$. Apply a vertex-hunting algorithm to the cloud $\{g_\theta(x_i)\}_{i=1}^n$ to estimate vertices $\hat V$, and decode each example by barycentric coordinates against $\hat V$. \newline
\textbf{(R2) Architectural bottleneck.} Parameterize $g_\theta(x)=\hat V\,\alpha_\theta(x)$ with $\hat V\in\R^{M\times K}$ column-stochastic and $\alpha_\theta(x)\in\Delta^{K-1}$, so that the simplex factorization is enforced at training time. The columns of $\hat V$ are the candidate vertices. The mixing matrix is then read off as $\hat\Pi = \hat V\diag(\hat c)$ with $\hat c$ solving $\hat V\hat c=\mathbf{1}_M$, mirroring Theorem~\ref{thm:identifiability}.\newline

The $M$-way classifier of the post-hoc fitting procedure is trained independently, and the specific fitter choice can improve the performance on real data. The architectural bottleneck provides a strong inductive bias that aids recovery with weak trunks (e.g., frozen backbones), and avoids potential failure modes of fitters, though it may be prone to local minima in $\hat V$ when using highly expressive models. We analyze empirical evidence in Fig.~\ref{fig:simplex_geometry}. Algorithm~\ref{alg:multiclass-cwola} summarizes the prior-free pipeline. \newline

\begin{algorithm}[t]
\caption{Multiclass CWoLa}
\label{alg:multiclass-cwola}
\begin{algorithmic}[1]
\Require Unlabeled mixtures $\D_1,\ldots,\D_M$, target class $K\le M$, recovery mode $\in\{\textsc{R1},\textsc{R2}\}$
\State Form the supervised dataset $\{(x_i,m_i)\}$ where $m_i$ is the mixture identity
\If{mode is \textsc{R1}}
    \State Train an $M$-way classifier $g_\theta:\X\to\Delta^{M-1}$ via cross-entropy
    \State Compute $u_i=g_\theta(x_i)$
    \State Fit $\hat V=\widehat{\mathcal{V}}(\{u_i\})$ via vertex hunting 
    \State Decode $\hat\alpha(x)=\arg\min_{\alpha\in\Delta^{K-1}}\|g_\theta(x)-\hat V\alpha\|_2^2$
\ElsIf{mode is \textsc{R2}}
    \State Train $g_\theta(x)=\hat V\,\alpha_\theta(x)$ end-to-end via cross-entropy with bottleneck warm-up, $\hat V$ column-stochastic
    \State Set $\hat\alpha(x)=\alpha_\theta(x)$
\EndIf
\State Recover $\hat\Pi=\hat V\diag(\hat c)$ with $\hat c$ the solution of $\hat V\hat c=\mathbf{1}_M$
\State \Return latent-class predictor $\hat\alpha$ (or $h_\phi$) and recovered mixing matrix $\hat\Pi$
\end{algorithmic}
\end{algorithm}

\textbf{Scope and Failure Modes.} The validity of multiclass CWoLa relies on our core assumptions. We identify three primary failure modes:
(i) If $\rank(\Pi) < K$, the simplex collapses to a dimension less than $K-1$.
(ii) If \ref{ass:irreducibility} is violated, the data support does not reach the vertices $v_k$ and $h^\star$ is not identifiable.
(iii) If class-conditionals $p_k$ vary across mixtures, \ref{ass:shared} fails, the shared factorization $V$ does not exist, and recovery is ill-posed. 

Our identifiability analysis relies on the anchor condition A3. In the square case $M=K$, the factorization $g_\theta = \hat{V}\alpha_\theta$ imposes no dimensionality reduction and matches the factorized network of VolMinNet \citep{Li2021ProvablyEL}, which achieves identifiability through an explicit volume regularizer under a weaker sufficiently scattered condition. The factorization acts as a genuine bottleneck only for $M > K$; whether sufficiently-scattered conditions extend to this rectangular regime is an open question.

\section{Experiments and analysis}

\subsection{Experimental setup}
\label{sec:experimental-setup}

We evaluate on MNIST~\citep{LeCun2005TheMD}, Fashion-MNIST~\citep{Xiao2017FashionMNISTAN}, CIFAR-10~\citep{krizhevsky2010cifar}, and Galaxy10 DECaLS~\citep{Lintott2008GalaxyZM, galaxyCALS}.
MNIST and Fashion-MNIST provide controlled grayscale benchmarks, CIFAR-10 serves as the main scaling benchmark, and Galaxy10 provides a real-world morphology dataset with $K=10$ classes.
All inputs are scaled to $[0,1]$. Galaxy10 images are resized to $224\times224$ and ImageNet-normalised inside the backbone \cite{Imagenet2009}.

For each experiment, we construct $M$ unlabeled mixture datasets from a labeled pool, hiding class labels during training. The mixing matrix $\Pi \in \mathbb{R}^{M\times K}$ is sampled row-wise as $\pi_m \sim \mathrm{Dirichlet}(\alpha \mathbf{1}_K)$, with $\alpha=0.8$ for MNIST, Fashion-MNIST, and CIFAR-10, and $\alpha=0.5$ for Galaxy10. When $M \ge K$, the Dirichlet draw is stabilized by blending each row with a fixed cyclic identity-like template before row renormalization (Appendix~\ref{app:experimental-details}). Each mixture $\mathcal{D}_m$ is then generated by sampling $k \sim \pi_m$ and drawing an example uniformly from the class-$k$ training pool. Only mixture identity $m$ is observed by prior-free methods; labels are used only for post-hoc alignment and evaluation.

For the purity study we use cyclic-purity mixtures. Each mixture $m$ is dominated by one class $c(m)=m \bmod K$, with $\pi_{m,c(m)}=\rho$, while the remaining mass is split uniformly across the other $K-1$ classes as $\pi_{m,k} = (1-\rho)/(K-1)$ for $k \neq c(m)$. Thus $\rho=1$ gives single-class mixtures and $\rho=1/K$ gives identical uniform mixtures.

For MNIST, Fashion-MNIST, and CIFAR-10, all methods use the same four-layer CNN with channel widths $(64,128,256,512)$ and a 512-dimensional embedding.
For Galaxy10, we use a frozen ImageNet-pretrained ResNet50 \cite{He2015DeepRL} trunk and train a projection head $(2048\to512)$ with ReLU.
Models are trained with AdamW, cosine annealing with linear warm-up, and early stopping on validation mixture accuracy.

The empirical analysis is organized around the main claims of the paper.
We first visualize the learned mixture posterior to test the predicted simplex geometry when the learned classifier $g_\theta(x)$ only approximates $g^\star(x)$.
We then compare prior-free design choices, including post-hoc simplex fitting, the architectural bottleneck, and vertex-spread regularization, on MNIST, Fashion-MNIST, and CIFAR-10.
CIFAR-10 is used for robustness comparisons against supervised, prior-aware, and other weakly-supervised prior-free baseline methods, measuring aligned latent accuracy across both $K=M$ scaling and fixed-$K$ scaling with varying $M$. In our context, accuracy is computed with respect to the mixture identity and so is informative while aligned accuracy corresponds to the accuracy with respect to the latent label.
For Galaxy10 purity experiments, we use a frozen ImageNet-pretrained ResNet50 trunk and train a projection head $(2048\to512)$ with ReLU.
The final Galaxy10 comparison in Table~\ref{tab:method_comparison} additionally reports a finetuned variant in which the backbone is unfrozen.

\subsection{Posterior simplex geometry and method analysis}
\label{sec:posterior-simplex-analysis}
\begin{figure}[h]
    \centering
    \includegraphics[width=\textwidth]{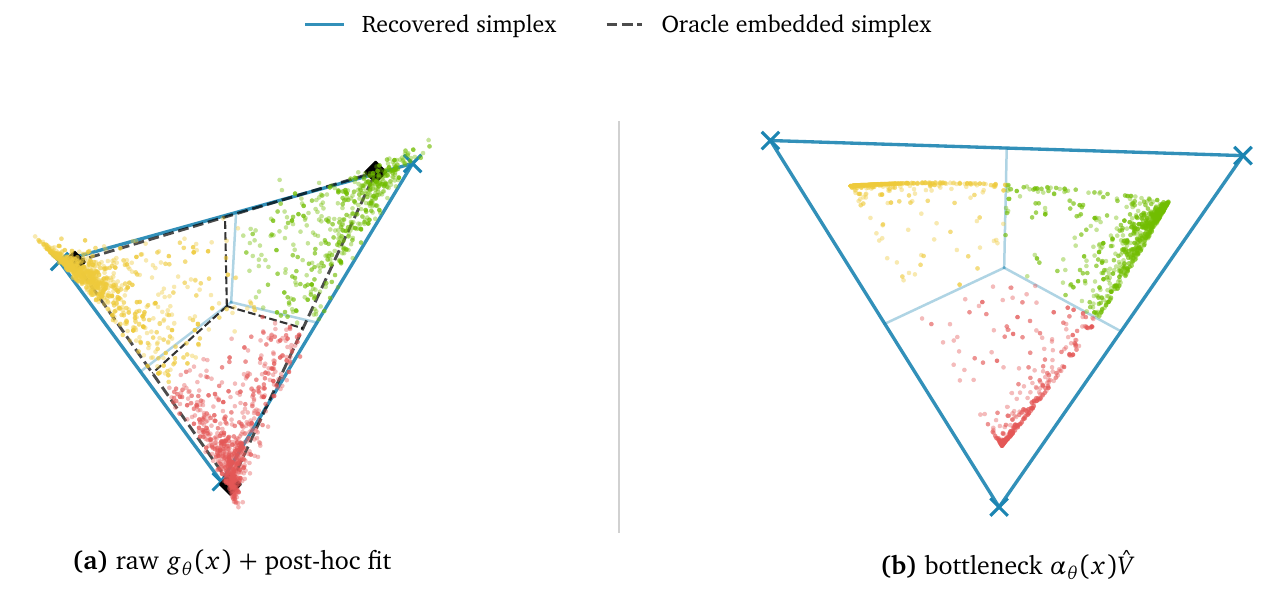}
    \caption{
    Posterior-simplex geometry in source-posterior space for CIFAR-10 with
    $K=3$ and $M=6$. Points are plotted in
    $\Delta^{M-1}$ and shown through a shared two-dimensional PCA projection.
    Colors indicate latent class structure for visualization only, not training
    supervision. In panel (a), the cloud is the raw source posterior
    $g_\theta(x)$ from a regular mixture classifier; the solid simplex is the
    post-hoc fitted simplex, and the dashed simplex is the oracle embedded
    simplex. In panel (b), points are bottleneck reconstructions
    $\alpha_\theta(x)\hat V$ and the solid
    simplex is the learned matrix from the bottleneck, namely the columns of $\hat{V}$.
    }
    \label{fig:simplex_geometry}
\end{figure}
We first examine whether the mixture posterior learned from mixture identities exhibits the simplex geometry predicted by the theory.
Figure~\ref{fig:simplex_geometry} visualizes the object characterized by the
posterior-simplex theorem: although the network is trained only to predict
mixture identity, its source-posterior outputs organize around a
$(K-1)$-dimensional simplex embedded in the larger $(M-1)$-dimensional
mixture-posterior space. The post-hoc method recovers this simplex after
training by fitting vertices to the source-posterior cloud, while the
bottleneck method imposes the same factorization during training and maps
examples through $\alpha_\theta(x)\hat V$. 

The plot supports the central geometric claim: although the model is trained only to predict mixture identity, the learned posteriors organise around a low-dimensional simplex whose vertices correspond to latent classes.

Most examples concentrate near one of the oracle vertices, while ambiguous or harder examples occupy the simplex edges and interior. The oracle simplex boundaries recover the dominant organization of the examples, with most points of each latent class concentrated in the corresponding simplex region. The remaining overlap is expected: the learned source classifier is only an empirical approximation to the Bayes-optimal posterior.
This is the empirical signature required by our demixing argument: latent classes are not observed directly, but they become recoverable as extreme points of the source-posterior geometry.

We then compare the two principled recovery methods we exposed in Sec.~\ref{sec:methodology} to exploit this geometry.
The first is a \emph{simplex} fitting approach: train a source classifier, treat its posterior cloud as fixed, and fit a simplex to recover the latent vertices.
This separates representation learning from geometric recovery and gives a direct diagnostic of whether the learned posterior has the predicted shape.
Fig.~\ref{fig:simplex_geometry}~\textbf{(a)} illustrates this strategy.
The fitted simplex closely tracks the oracle simplex, indicating that the class vertices can be recovered from the posterior cloud without using labels during training. 

The second approach is a \emph{bottleneck} approach: instead of fitting a simplex after training, the model is constrained architecturally so that source prediction factors through a $K$-dimensional latent bottleneck.
This encourages the latent representation itself to align with the simplex vertices. 
Fig.~\ref{fig:simplex_geometry}~\textbf{(b)} shows that the bottleneck head also produces vertex-concentrated posteriors aligned with the oracle simplex. Because of regularization, entropy penalties, slack variables, temperature smoothing, or constraints that make the factorization numerically stable, the bottleneck is discouraged to predict extreme one-hot $\alpha_\theta(x)$. Explaining why the data cloud stays inside the learned simplex instead of reaching the corners.

Thus, both approaches use the same theoretical object---the posterior simplex---but in different ways: post-hoc fitting recovers the simplex after learning, while the bottleneck head builds the simplex structure into the learning problem. 

We next test whether prior-free recovery is sensitive to the choice of simplex extraction method.
The fitter-robustness analysis in Fig.~\ref{fig:fitter-robustness}
and Fig.~\ref{fig:heatmap-fitters} shows that simplex recovery is not
an artifact of a particular post-hoc fitting rule. Across datasets and
mixture regimes, several fitters recover simplexes close to the known-prior
demixing oracle, supporting the geometric interpretation that the posterior
cloud contains a stable signal of the hidden mixing structure. This matters
beyond aligned classification accuracy: stable vertex recovery also
allows latent class proportions estimation within mixtures, which may be
the primary scientific target in composition-focused applications.

\subsection{Robustness and baseline comparisons}
\begin{figure}[h]
    \centering
    \includegraphics[width=\textwidth]{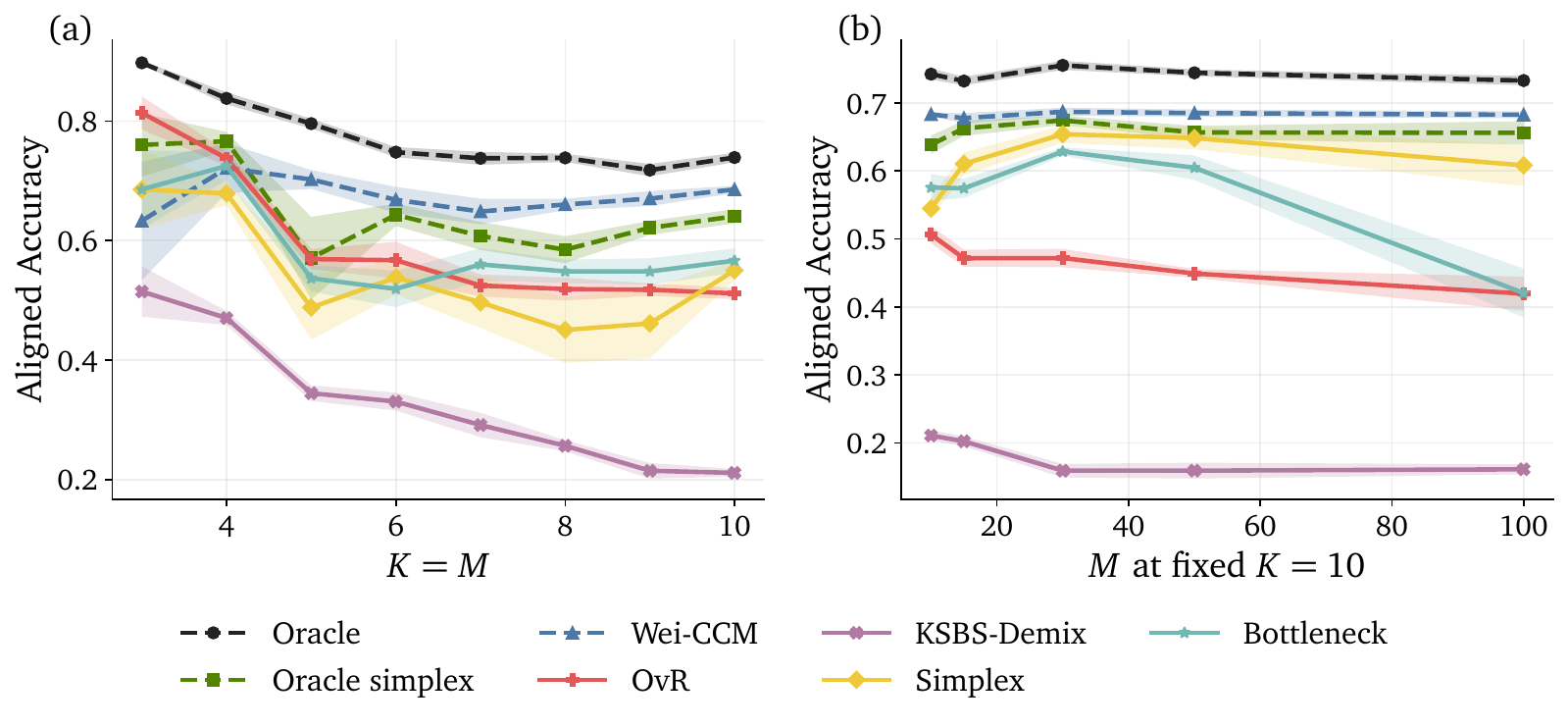}
    \caption{
    Performance as a function of mixture complexity on CIFAR-10. Left: aligned accuracy versus the balanced setting K=M, where the number of latent classes and mixtures increase jointly. Right: aligned accuracy versus the number of mixtures M at fixed K=10. Shaded regions denote one standard error across random seeds. Solid lines represent methods that receives the same amount of information as our methods.
    }
    \label{fig:robustness-cifar}
\end{figure}
We next compare our prior-free methods to existing baselines as the number of classes and mixtures increases.
Fig.~\ref{fig:robustness-cifar} reports aligned accuracy on CIFAR-10 in two regimes: balanced scaling with $K=M$, and fixed $K=10$ with increasing $M$.
The supervised oracle gives an upper reference, while the oracle simplex and Wei-CCM/RCM baselines assume access to mixture proportions~\citep{Tang2022MulticlassCF,wei2024consistent}.
Among methods without class priors, OvR is the main fair reference~\citep{Rifkin2004InDO}, together with the prior-free \emph{KSBS-Demix} baseline~\citep{KatzSamuels2017DecontaminationOM}. However, OvR comes at the heavy computational cost of having to train $M-1$ classifiers for $M$ mixtures.

Across both regimes, our methods are substantially more stable than the prior-free demixing baseline.
In the balanced setting $K=M$, all weakly supervised methods degrade as the number of classes grows, but the simplex and bottleneck variants remain close to the known-prior simplex oracle reference and overtake OvR in the harder high-$K$ regimes.
At fixed $K=10$, increasing the number of mixtures improves the geometry fitting: the simplex variant rises from $0.55$ at $M=10$ to about $0.65$ for $M\in\{30,50\}$, nearly matching the known-prior simplex oracle.
This supports the geometric interpretation that additional mixtures provide redundant constraints for estimating the latent simplex.

The main gap that remains is to the fully supervised oracle, which uses labels unavailable to all weakly supervised methods.
By contrast, the gap to the known-prior simplex is small in the overcomplete regime, showing that most of the performance loss comes from not observing labels, rather than from estimating the simplex without priors.
\begin{figure}[h]
    \centering
    \includegraphics[width=1\textwidth]{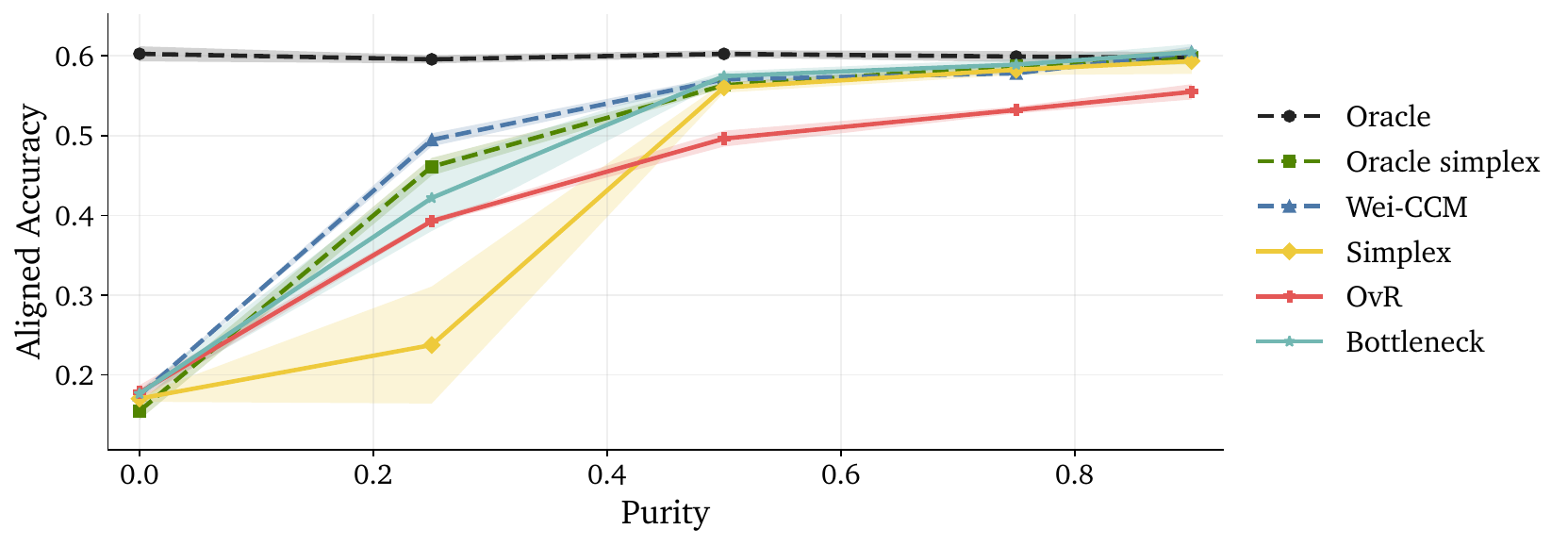}
    \caption{
    Robustness to mixture purity on Galaxy10 DECaLS.
    We construct cyclic-purity mixtures with $K=M=10$ while keeping the image data, backbone, and augmentation profile fixed.
    Low purity corresponds to weak class enrichment and approaches the unidentifiable regime; high purity gives mixtures with clearer anchor structure.
    Accuracy increases with purity for all mixture-based methods.
    Our simplex and bottleneck variants outperform the no-prior OvR baseline at moderate and high purity, while approaching the known-prior simplex and Wei-CCM references.
    }
    \label{fig:purity}
\end{figure}

Finally, we set ourselves in a real data context. For that, we tested how the method behaves when the mixtures themselves become less informative using real data. First, we confirm similar robustness across $M$ at fixed $K=10$ as with the other dataset (see Fig.~\ref{fig:gal_m_sweep}).
Then in applications, constructing mixtures with strong class-specific enrichment may be difficult, so the relevant question is not only whether demixing works under ideal mixtures, but how gracefully it degrades as mixture purity decreases.
We study this on Galaxy10 using real astronomical images and controlled cyclic-purity mixtures with $K=M=10$.

To avoid tuning specifically for this diagnostic, we use the same fixed ImageNet-pretrained ResNet50 backbone and symmetry-only augmentation profile throughout.
Thus, the experiment isolates the effect of mixture quality rather than improvements from task-specific finetuning or stronger augmentation.
Fig.~\ref{fig:purity} shows the expected behaviour: when purity is near zero, the mixtures carry little class-specific information and all weakly supervised methods approach the unidentifiable regime.
As purity increases, accuracy improves monotonically and our bottleneck/simplex variants approach the known-prior simplex and Wei-CCM references. Similar trends are confirmed on CIFAR-10 in Fig.~\ref{fig:purity_cifar}.

The comparison to OvR is especially informative.
OvR is a fair no-prior baseline but is more costly because it requires solving multiple binary reductions.
At moderate and high purity, our methods open a clear gap over OvR while remaining close to the oracle-prior methods.
This suggests that the posterior-simplex structure is not only visible in controlled image benchmarks, but also provides a useful and computationally lighter recovery mechanism on real astronomical data.
\begin{table*}[h]
\centering
\begin{tblr}{
  width = \textwidth,
  colspec = {X[l,m]X[c]X[c]X[c]}, 
  colsep = 18pt,
  stretch = 1.0,
  column{1} = {leftsep=0pt},
  column{4} = {rightsep=0pt},
}
\toprule
Technique & Method & Accuracy & ECE \\
\midrule
Supervised
& Oracle
& $0.802 \pm 0.023$
& $0.032 \pm 0.004$ \\

\midrule[dotted]

& Wei CCM
& $0.686 \pm 0.015$
& $0.069 \pm 0.046$ \\
Prior-aware  & Wei RCM
& $0.577 \pm 0.013$
& $0.371 \pm 0.020$ \\
& Oracle Simplex
& $0.661 \pm 0.012$
& $0.116 \pm 0.025$ \\
\midrule[dotted]

& OvR
& $0.476 \pm 0.034$
& $0.272 \pm 0.034$ \\
Prior-free 
& \textbf{Bottleneck}
& $\mathbf{0.503 \pm 0.021}$
& $\mathbf{0.197 \pm 0.052}$ \\
& \textbf{Simplex}
& $\mathbf{0.584 \pm 0.051}$
& $\mathbf{0.123 \pm 0.031}$ \\

\bottomrule
\end{tblr}

\caption{
Finetuned Galaxy10 comparison at $K=10$ and $M=20$.
Results are reported as mean $\pm$ standard deviation over three random seeds.
Accuracy is permutation-aligned latent accuracy; ECE is expected calibration error, where lower is better.
Oracle is fully supervised.
Wei CCM and Wei RCM assume access to class-prior information, and Oracle Simplex uses the true mixture proportions.
OvR, Bottleneck, and Simplex are prior-free.
Simplex is the strongest prior-free method, improving substantially over OvR in both accuracy and calibration.
}
\label{tab:method_comparison}
\end{table*}

As a final real-data evaluation, we ask whether the same prior-free recovery remains competitive when the representation is allowed to adapt to Galaxy10.
Unlike the purity diagnostic above, which keeps the ResNet backbone fixed to isolate the effect of mixture quality, this experiment uses $180^\circ$-rotations and horizontal/vertical flips as data augmentation, and unfreezes the backbone model weights.
Table~\ref{tab:method_comparison} reports both aligned accuracy and calibration. The same study on frozen model can be found in Table~\ref{tab:method_comparison_notune}.

Among prior-free methods, Simplex fitting performs best, improving over OvR by $10.8$ accuracy points and substantially reducing calibration error.
The bottleneck variant is also competitive with OvR in accuracy, but the post-hoc Simplex fitter gives the strongest overall improvement.
Prior-aware methods remain stronger, as expected, because they use additional class-prior information unavailable to fully prior-free methods.
Overall, the result shows that the posterior-simplex recovery remains effective beyond the frozen-backbone diagnostic and gives a strong prior-free baseline on real astronomical images.

So far, the implementations implicitly assume that the exact number of true latent classes $K$ is known a priori. However, our simplex fitting framework does not require this.
Because the direct $M$-way classification pipeline only introduces the target class count $K$ inside the post-hoc simplex fitter, $K$ can be treated as a discoverable quantity once the calibrated source posteriors are in hand. In Appendix~\ref{app:true_K_Discovery} we demonstrate this empirically for the Galaxy10 dataset.

\section{Discussion and conclusion}
\label{sec:discussion}

We have shown that multiclass classification without instance-level labels admits a geometric solution when supervision is limited to mixture identity. Under standard mixture assumptions, the Bayes-optimal mixture posterior forms a latent simplex whose vertices identify both the class posterior and the hidden mixing matrix, enabling recovery of instance-level predictions as well as mixture-level class proportions. We instantiate this idea with two prior-free procedures: post-hoc vertex hunting and a bottleneck architecture enforcing $g_\theta=\hat V\alpha_\theta$. Experiments on MNIST, Fashion-MNIST, CIFAR-10, and Galaxy10 DECaLS show that these methods approach the oracle simplex baseline, especially in overcomplete-mixture regimes.

Unlike binary CWoLa~\citep{Metodiev:2017vrx}, which separates two classes, multiclass CWoLa can separate $K\leq M$ classes. Both LLP and prior multiclass mixture-classification methods \citep{Tang2022MulticlassCF,wei2024consistent}, require known class proportions while for our approach mixture identity alone suffices alongside the recovered geometry of $g_\theta$. 

The main limitations are that weak anchors shrink the simplex and degrade identification, while violations of the shared-class-conditionals assumption \asmp{ass:shared} can make recovered classes reflect mixture-specific artifacts rather than semantic categories. In practice, recovered classes should
therefore be validated on held-out labels, especially before high-stakes use.

In many real-world applications, data are collected under heterogeneous conditions that naturally produce mixtures with different latent class compositions. By reducing dependence on exhaustive instance-level annotation, weakly-supervised methods may lower the cost of multiclass learning in domains requiring scarce expert labour, such as medicine, astronomy, and particle physics, and may reduce the human burden of large-scale annotation pipelines. This benefit does not remove the need for expert validation, calibration,
and auditing.

Future work includes relaxing anchor assumptions, studying how biased mixtures that violate the shared class-conditionals assumption can be integrated, and extending the framework to partial or noisy labels.

\begin{ack}
The idea of exploring a multiclass extension of CWoLa originated with J. I.-N. during his exchange in Tobias Golling's group, following inspiring discussions with Jesse Thaler during his concurrent visit to the Geneva group. We are grateful to Tobias and his group for the warm hospitality and to both Jesse and Tobias for valuable insights during this period.

This work was supported by the European Union's Horizon Europe research and innovation programme under the Marie Sk\l{}odowska-Curie grant agreement No. 101168829, Challenging AI with Challenges from Physics: How to solve fundamental problems in Physics by AI and vice versa (AIPHY). We thank the University of Copenhagen's SCIENCE AI Centre for access to the centre's GPU cluster.

\end{ack}

\bibliographystyle{plainnat}
\bibliography{ref}

@article{lecun:hal-04206682,
  TITLE = {{Deep learning}},
  AUTHOR = {Lecun, Yann and Bengio, Yoshua and Hinton, Geoffrey},
  URL = {https://hal.science/hal-04206682},
  JOURNAL = {{Nature}},
  PUBLISHER = {{Nature Publishing Group}},
  VOLUME = {521},
  NUMBER = {7553},
  PAGES = {436-444},
  YEAR = {2015},
  MONTH = May,
  DOI = {10.1038/nature14539},
  HAL_ID = {hal-04206682},
  HAL_VERSION = {v1},
}

@misc{LeCun2005TheMD,
  title={The mnist database of handwritten digits},
  author={Yann LeCun and Corinna Cortes},
  year={2005},
  url={https://api.semanticscholar.org/CorpusID:60282629}
}

@article{krizhevsky2010cifar,
  title={Cifar-10 (canadian institute for advanced research)},
  author={Krizhevsky, Alex and Nair, Vinod and Hinton, Geoffrey},
  journal={URL http://www.cs.toronto.edu/kriz/cifar. html},
  volume={5},
  number={4},
  pages={1},
  year={2010}
}

@article{Xiao2017FashionMNISTAN,
  title={Fashion-MNIST: a Novel Image Dataset for Benchmarking Machine Learning Algorithms},
  author={Han Xiao and Kashif Rasul and Roland Vollgraf},
  journal={ArXiv},
  year={2017},
  volume={abs/1708.07747},
  url={https://api.semanticscholar.org/CorpusID:702279}
}

@article{Lintott2008GalaxyZM,
  title={Galaxy Zoo: morphologies derived from visual inspection of galaxies from the Sloan Digital Sky Survey},
  author={Chris J. Lintott and Kevin Schawinski and An{\v{z}}e Slosar and Kate R. Land and Steven Bamford and Daniel Thomas and M. Jordan Raddick and Robert C. Nichol and Alexander S. Szalay and Daniel Andreescu and P. G. Murray and Jan van den Berg},
  journal={Monthly Notices of the Royal Astronomical Society},
  year={2008},
  volume={389},
  pages={1179-1189},
  url={https://api.semanticscholar.org/CorpusID:15279243}
}

@article{galaxyCALS,
    author = {Walmsley, Mike and Lintott, Chris and G{\'e}ron, Tobias and Kruk, Sandor and Krawczyk, Coleman and Willett, Kyle W and Bamford, Steven and Kelvin, Lee S and Fortson, Lucy and Gal, Yarin and Keel, William and Masters, Karen L and Mehta, Vihang and Simmons, Brooke D and Smethurst, Rebecca and Smith, Lewis and Baeten, Elisabeth M and Macmillan, Christine},
    title = {Galaxy Zoo DECaLS: Detailed visual morphology measurements from volunteers and deep learning for 314\,000 galaxies},
    journal = {Monthly Notices of the Royal Astronomical Society},
    volume = {509},
    number = {3},
    pages = {3966-3988},
    year = {2022},
    month = {01},
    issn = {0035-8711},
    doi = {10.1093/mnras/stab2093},
    url = {https://doi.org/10.1093/mnras/stab2093},
    eprint = {https://academic.oup.com/mnras/article-pdf/509/3/3966/45718280/stab2093.pdf},
}

@article{He2015DeepRL,
  title={Deep Residual Learning for Image Recognition},
  author={Kaiming He and X. Zhang and Shaoqing Ren and Jian Sun},
  journal={2016 IEEE Conference on Computer Vision and Pattern Recognition (CVPR)},
  year={2015},
  pages={770-778},
  url={https://api.semanticscholar.org/CorpusID:206594692}
}

@INPROCEEDINGS{Imagenet2009,
  author={Deng, Jia and Dong, Wei and Socher, Richard and Li, Li-Jia and Kai Li and Li Fei-Fei},
  booktitle={2009 IEEE Conference on Computer Vision and Pattern Recognition}, 
  title={ImageNet: A large-scale hierarchical image database}, 
  year={2009},
  volume={},
  number={},
  pages={248-255},
  doi={10.1109/CVPR.2009.5206848}}

@article{Russakovsky2014ImageNetLS,
  title={ImageNet Large Scale Visual Recognition Challenge},
  author={Olga Russakovsky and Jia Deng and Hao Su and Jonathan Krause and Sanjeev Satheesh and Sean Ma and Zhiheng Huang and Andrej Karpathy and Aditya Khosla and Michael S. Bernstein and Alexander C. Berg and Li Fei-Fei},
  journal={International Journal of Computer Vision},
  year={2014},
  volume={115},
  pages={211 - 252},
  url={https://api.semanticscholar.org/CorpusID:2930547}
}

@article{Ratner2017SnorkelRT,
  title={Snorkel: Rapid Training Data Creation with Weak Supervision},
  author={Alexander J. Ratner and Stephen H. Bach and Henry R. Ehrenberg and Jason Alan Fries and Sen Wu and Christopher R{\'e}},
  journal={Proceedings of the VLDB Endowment. International Conference on Very Large Data Bases},
  year={2017},
  volume={11 3},
  pages={
          269-282
        },
  url={https://api.semanticscholar.org/CorpusID:6730236}
}

@article{Darg:2009rc,
    author = "Darg, D. W. and others",
    title = "{Galaxy Zoo: the fraction of merging galaxies in the SDSS and their morphologies}",
    eprint = "0903.4937",
    archivePrefix = "arXiv",
    primaryClass = "astro-ph.GA",
    doi = "10.1111/j.1365-2966.2009.15686.x",
    journal = "Mon. Not. Roy. Astron. Soc.",
    volume = "401",
    pages = "1043",
    year = "2010"
}

@article{Wang2020AnnotationefficientDL,
  title={Annotation-efficient deep learning for automatic medical image segmentation},
  author={Shanshan Wang and Cheng Li and Rongpin Wang and Zaiyi Liu and Meiyun Wang and Hongna Tan and Yaping Wu and Xinfeng Liu and Hui Sun and Rui Yang and Xin Liu and Jie Chen and Hui-Chong Zhou and Ismail Ben Ayed and Hairong Zheng},
  journal={Nature Communications},
  year={2020},
  volume={12},
  url={https://api.semanticscholar.org/CorpusID:237604951}
}

@article{Chen2020ASF,
  title={A Simple Framework for Contrastive Learning of Visual Representations},
  author={Ting Chen and Simon Kornblith and Mohammad Norouzi and Geoffrey E. Hinton},
  journal={ArXiv},
  year={2020},
  volume={abs/2002.05709},
  url={https://api.semanticscholar.org/CorpusID:211096730}
}

@article{He2021MaskedAA,
  title={Masked Autoencoders Are Scalable Vision Learners},
  author={Kaiming He and Xinlei Chen and Saining Xie and Yanghao Li and Piotr Doll'ar and Ross B. Girshick},
  journal={2022 IEEE/CVF Conference on Computer Vision and Pattern Recognition (CVPR)},
  year={2021},
  pages={15979-15988},
  url={https://api.semanticscholar.org/CorpusID:243985980}
}

@inproceedings{Han2018CoteachingRT,
  title={Co-teaching: Robust training of deep neural networks with extremely noisy labels},
  author={Bo Han and Quanming Yao and Xingrui Yu and Gang Niu and Miao Xu and Weihua Hu and Ivor Wai-Hung Tsang and Masashi Sugiyama},
  booktitle={Neural Information Processing Systems},
  year={2018},
  url={https://api.semanticscholar.org/CorpusID:52065462}
}

@article{Metodiev:2017vrx,
    author = "Metodiev, Eric M. and Nachman, Benjamin and Thaler, Jesse",
    title = "{Classification without labels: Learning from mixed samples in high energy physics}",
    eprint = "1708.02949",
    archivePrefix = "arXiv",
    primaryClass = "hep-ph",
    reportNumber = "MIT--CTP-4922",
    doi = "10.1007/JHEP10(2017)174",
    journal = "JHEP",
    volume = "10",
    pages = "174",
    year = "2017"
}

@article{Komiske2018LearningTC,
  title={Learning to classify from impure samples with high-dimensional data},
  author={Patrick T. Komiske and Eric M. Metodiev and Benjamin Philip Nachman and Matthew D. Schwartz},
  journal={Physical Review D},
  year={2018},
  url={https://api.semanticscholar.org/CorpusID:54019676}
}

@article{Collins2018CWoLaHE,
  title={CWoLa Hunting: Extending the Bump Hunt with Machine Learning},
  author={Jack H. Collins and Kiel Howe and Benjamin Philip Nachman},
  journal={arXiv: High Energy Physics - Phenomenology},
  year={2018},
  url={https://api.semanticscholar.org/CorpusID:126361005}
}

@article{Rifkin2004InDO,
  title={In Defense of One-Vs-All Classification},
  author={Ryan M. Rifkin and Aldebaro Klautau},
  journal={J. Mach. Learn. Res.},
  year={2004},
  volume={5},
  pages={101-141},
  url={https://api.semanticscholar.org/CorpusID:13391792}
}

@inproceedings{Scott2013ClassificationWA,
  author    = {Scott, Clayton and Blanchard, Gilles and Handy, Gregory},
  title     = {Classification with Asymmetric Label Noise: Consistency and Maximal Denoising},
  booktitle = {Proceedings of the 26th Annual Conference on Learning Theory},
  pages     = {489--511},
  year      = {2013},
  editor    = {Shalev-Shwartz, Shai and Steinwart, Ingo},
  volume    = {30},
  series    = {Proceedings of Machine Learning Research},
  publisher = {PMLR},
  url       = {https://proceedings.mlr.press/v30/Scott13.html}
}

@inproceedings{Tang2022MulticlassCF,
  title={Multi-class Classification from Multiple Unlabeled Datasets with Partial Risk Regularization},
  author={Yuting Tang and Nan Lu and Tianyi Zhang and Masashi Sugiyama},
  booktitle={Asian Conference on Machine Learning},
  year={2022},
  url={https://api.semanticscholar.org/CorpusID:252918114}
}

@inproceedings{
wei2024consistent,
title={Consistent Multi-Class Classification from Multiple Unlabeled Datasets},
author={Zixi Wei and Senlin Shu and Yuzhou Cao and Hongxin Wei and Bo An and Lei Feng},
booktitle={The Twelfth International Conference on Learning Representations},
year={2024},
url={https://openreview.net/forum?id=fW7DOHDQvF}
}

@article{Scott2020LearningFL,
  title={Learning from Label Proportions: A Mutual Contamination Framework},
  author={Clayton Scott and Jianxin Zhang},
  journal={ArXiv},
  year={2020},
  volume={abs/2006.07330},
  url={https://api.semanticscholar.org/CorpusID:263786144}
}

@article{KatzSamuels2017DecontaminationOM,
  title={Decontamination of Mutual Contamination Models},
  author={Julian Katz-Samuels and Gilles Blanchard and Clayton D. Scott},
  journal={J. Mach. Learn. Res.},
  year={2017},
  volume={20},
  pages={41:1-41:57},
  url={https://api.semanticscholar.org/CorpusID:88515763}
}

@inproceedings{Donoho2003:nmf,
author = {Donoho, David and Stodden, Victoria},
title = {When does non-negative matrix factorization give a correct decomposition into parts?},
year = {2003},
publisher = {MIT Press},
address = {Cambridge, MA, USA},
booktitle = {Proceedings of the 17th International Conference on Neural Information Processing Systems},
pages = {1141--1148},
numpages = {8},
location = {Whistler, British Columbia, Canada},
series = {NIPS'03}
}

@article{Arora2011ComputingAN,
  title={Computing a nonnegative matrix factorization -- provably},
  author={Sanjeev Arora and Rong Ge and Ravi Kannan and Ankur Moitra},
  journal={ArXiv},
  year={2011},
  volume={abs/1111.0952},
  url={https://api.semanticscholar.org/CorpusID:15579971}
}

@article{Arora2012LearningTM,
  title={Learning Topic Models -- Going beyond SVD},
  author={Sanjeev Arora and Rong Ge and Ankur Moitra},
  journal={2012 IEEE 53rd Annual Symposium on Foundations of Computer Science},
  year={2012},
  pages={1-10},
  url={https://api.semanticscholar.org/CorpusID:10004443}
}

@article{Arora2012APA,
  title={A Practical Algorithm for Topic Modeling with Provable Guarantees},
  author={Sanjeev Arora and Rong Ge and Yoni Halpern and David Mimno and Ankur Moitra and David A. Sontag and Yichen Wu and Michael Zhu},
  journal={ArXiv},
  year={2012},
  volume={abs/1212.4777},
  url={https://api.semanticscholar.org/CorpusID:9220219}
}

@article{Huang2016AnchorFreeCT,
  title={Anchor-Free Correlated Topic Modeling: Identifiability and Algorithm},
  author={Kejun Huang and Xiao Fu and N. Sidiropoulos},
  journal={ArXiv},
  year={2016},
  volume={abs/1611.05010},
  url={https://api.semanticscholar.org/CorpusID:5255990}
}

@article{BioucasDias2012HyperspectralUO,
  title={Hyperspectral Unmixing Overview: Geometrical, Statistical, and Sparse Regression-Based Approaches},
  author={Jos{\'e} M. Bioucas-Dias and Antonio J. Plaza and Nicolas Dobigeon and Mario Parente and Qian Du and Paul D. Gader and Jocelyn Chanussot},
  journal={IEEE Journal of Selected Topics in Applied Earth Observations and Remote Sensing},
  year={2012},
  volume={5},
  pages={354-379},
  url={https://api.semanticscholar.org/CorpusID:11112426}
}

@article{Ke2017UsingSF,
  title={Using SVD for Topic Modeling},
  author={Zheng Tracy Ke and Minzhe Wang},
  journal={Journal of the American Statistical Association},
  year={2017},
  volume={119},
  pages={434 - 449},
  url={https://api.semanticscholar.org/CorpusID:251928821}
}

@article{cutler1994archetypal,
 ISSN = {00401706},
 URL = {http://www.jstor.org/stable/1269949},
 author = {Adele Cutler and Leo Breiman},
 journal = {Technometrics},
 number = {4},
 pages = {338--347},
 publisher = {[Taylor & Francis, Ltd., American Statistical Association, American Society for Quality]},
 title = {Archetypal Analysis},
 urldate = {2026-05-06},
 volume = {36},
 year = {1994}
}

@article{Chan2009ASF,
  author       = {Jesus Tordesillas and
                  Jonathan P. How},
  title        = {{MINVO} Basis: Finding Simplexes with Minimum Volume Enclosing Polynomial
                  Curves},
  journal      = {CoRR},
  volume       = {abs/2010.10726},
  year         = {2020},
  url          = {https://arxiv.org/abs/2010.10726},
  eprinttype   = {arXiv},
  eprint       = {2010.10726},
  bibsource    = {dblp computer science bibliography, https://dblp.org}
}

@article{Liu2015,
author = {Liu, Tongliang and Tao, Dacheng},
title = {Classification with Noisy Labels by Importance Reweighting},
year = {2016},
issue_date = {March 2016},
publisher = {IEEE Computer Society},
address = {USA},
volume = {38},
number = {3},
issn = {0162-8828},
url = {https://doi.org/10.1109/TPAMI.2015.2456899},
doi = {10.1109/TPAMI.2015.2456899},
journal = {IEEE Trans. Pattern Anal. Mach. Intell.},
month = mar,
pages = {447--461},
numpages = {15}
}

@inproceedings{Zhang2021LearningNT,
  title={Learning Noise Transition Matrix from Only Noisy Labels via Total Variation Regularization},
  author={Yivan Zhang and Gang Niu and Masashi Sugiyama},
  booktitle={International Conference on Machine Learning},
  year={2021},
  url={https://api.semanticscholar.org/CorpusID:231802334}
}

@article{Li2021ProvablyEL,
  title={Provably End-to-end Label-Noise Learning without Anchor Points},
  author={Xuefeng Li and Tongliang Liu and Bo Han and Gang Niu and Masashi Sugiyama},
  journal={ArXiv},
  year={2021},
  volume={abs/2102.02400},
  url={https://api.semanticscholar.org/CorpusID:231802306}
}

@misc{deLaFuente:SimplexDemixing,
  author = {{de la Fuente}, Gregorio and Thaler, Jesse},
  title  = {Simplex Demixing: Disentangling Multiple Light-Flavor Jets at Colliders},
  year   = {2026},
  note   = {to appear}
}


\appendix

\section{Technical Details for Experiments}

\textbf{Experiments Statistical Significance.} All error bars reported in the paper are the 1$\sigma$ standard deviations calculated over 3 independent runs with different random seeds. We assume approximately normally distributed errors for the reported metrics.

\textbf{Compute Resources.} All experiments for MNIST were run on local CPUs. FashionMNIST, CIFAR-10, and Galaxy10 DECaLS were executed on a shared internal cluster using a single A100 GPU with 40GB VRAM. Each individual experimental run completed in less than 6 hours. Total compute for the project, including preliminary and failed experiments, is estimated at less than 100 GPU-hours.

\textbf{Code Availability.} We release \texttt{MultiCWoLa}, a reusable library implementing the post-hoc and bottleneck approaches introduced in this work. The library is designed to facilitate the transfer and integration of these methods into new datasets, models, and application domains, and is publicly available here \href{https://github.com/rbonnetguerrini/MultiCWoLa}{\texttt{MultiCWoLa repository}}.

\textbf{LLM Resources.} The authors acknowledge the use of the paid Claude Code Pro and GitHub Copilot for code implementation, debugging, and initial drafting/editing of ideas derived by the authors.

\textbf{Dataset Licenses and Assets.} The datasets and pre-trained models used in this study are all publicly available.

%
%
%

\section{Proofs for Sec.~\ref{sec:methodology}}
\label{app:proofs}

This appendix collects the proofs of the three theorems of Sec.~\ref{sec:methodology}. Throughout, $c_k := \sum_{m=1}^M \pi_{mk}$ is the class abundance, $V := [v_1,\dots,v_K] \in \R^{M\times K}$ is the vertex matrix with $v_k = \Pi_{:,k}/c_k$, and $\bar h^\star$ is the latent posterior under the effective prior $\bar\pi_k = c_k/M$.

\subsection{Proof of Theorem~\ref{thm:simplex} (Posterior Simplex Theorem)}
\label{app:proofs:simplex}

\begin{proof}
For any $x \in \X$ the optimal mixture posterior is given under \asmp{ass:uniform} via Bayes' rule by 
\begin{equation}
g_m^\star(x) \;=\; \Pp(S=m \mid X=x) \;=\; \frac{q_m(x)\,\Pp(S=m)}{\sum_{j=1}^M q_j(x)\,\Pp(S=j)} \;=\; \frac{q_m(x)}{\sum_{j=1}^M q_j(x)}.
\end{equation}
By Eq.~\ref{eq:mixture-model} and \asmp{ass:shared} (which makes $p_k$ independent of $m$),
\begin{equation}
g_m^\star(x) \;=\; \frac{\sum_{k=1}^K \pi_{mk}\, p_k(x)}{\sum_{j=1}^M \sum_{k=1}^K \pi_{jk}\, p_k(x)} \;=\; \frac{\sum_{k=1}^K \pi_{mk}\, p_k(x)}{\sum_{k=1}^K c_k\, p_k(x)},
\end{equation}
where we use $\sum_{j=1}^M \pi_{jk} = c_k$.
To show that the mixture posterior lives on a convex hull we define $\alpha_k(x) := c_k p_k(x) / \sum_\ell c_\ell p_\ell(x)$. Then $\alpha_k(x) \ge 0$, $\sum_k \alpha_k(x) = 1$, and using $(v_k)_m = \pi_{mk}/c_k$,
\begin{equation}
\sum_{k=1}^K \alpha_k(x)\,(v_k)_m \;=\; \sum_{k=1}^K \frac{c_k\, p_k(x)}{\sum_\ell c_\ell\, p_\ell(x)} \cdot \frac{\pi_{mk}}{c_k} \;=\; \frac{\sum_{k=1}^K \pi_{mk}\, p_k(x)}{\sum_\ell c_\ell\, p_\ell(x)} \;=\; g_m^\star(x).
\end{equation}
Hence $g^\star(x) = \sum_k \alpha_k(x) v_k = V\alpha(x)$ is constrained to a convex hull.\newline
To show that $g^\star(x)$ is constrained to a $(K-1)$ simplex we note that each $v_k = \Pi_{:,k}/c_k$ is a positive rescaling of the $k$-th column of $\Pi$. Suppose $\sum_k \beta_k v_k = 0$ with $\sum_k \beta_k = 0$. Then $\sum_k (\beta_k/c_k)\,\Pi_{:,k} = 0$, and \asmp{ass:rank} forces $\beta_k/c_k = 0$ for all $k$, i.e.\ $\beta_k = 0$. Thus $v_1,\dots,v_K$ are affinely independent in $\R^M$, and $\conv\{v_1,\dots,v_K\}$ is a $(K-1)$-simplex.
Vertices lie in $\Delta^{M-1}$. Each $v_k$ has nonnegative entries and $\sum_m (v_k)_m = c_k^{-1} \sum_m \pi_{mk} = 1$. So $v_k \in \Delta^{M-1}$ and $\mathcal{S} \subset \Delta^{M-1}$.
\end{proof}

\subsection{Proof of Theorem~\ref{thm:sufficient-statistic} (Multiclass CWoLa Optimality)}
\label{app:proofs:sufficient}

\begin{proof} 
Let $\bar h^\star_k(x) := \alpha_k(x)$ and thus $g^\star(x)=V~\bar h^\star(x)$. Under uniform mixture sampling, the marginal density factorizes as
\begin{equation}
\mu_X(x) \;=\; \frac{1}{M}\sum_{m=1}^M q_m(x) \;=\; \frac{1}{M}\sum_{k=1}^K c_k\, p_k(x) \;=\; \sum_{k=1}^K \bar\pi_k\, p_k(x),
\end{equation}
with $\bar\pi_k = c_k/M$. Bayes' rule under this prior gives
\begin{equation}
    \Pp_{\bar\pi}(Y=k \mid X=x) = \frac{\bar\pi_k\, p_k(x)}{\sum_\ell \bar\pi_\ell\, p_\ell(x)} = \bar h^\star_k(x),    
\end{equation}
So $\bar h^\star$ is the latent posterior under the effective prior $\bar\pi$.

From Theorem~\ref{thm:simplex}, $g^\star(x) = V\bar h^\star(x)$. Since $V = \Pi C^{-1}$ with $C = \diag(c_1,\dots,c_K)$ positive definite and $\rank(\Pi) = K$ (\asmp{ass:rank}), we have $\rank(V) = K$. Hence $V^\top V$ is positive definite and the Moore--Penrose left inverse $V^+ = (V^\top V)^{-1} V^\top$ satisfies $V^+ V = I_K$, giving
\begin{equation}
\bar h^\star(x) \;=\; V^+ g^\star(x) \;=\; (V^\top V)^{-1} V^\top g^\star(x).
\end{equation}
For the actual class prior $\pi^Y_k := \Pp(Y=k)$, Bayes' rule gives $h^\star_k(x) = \pi^Y_k\, p_k(x) / \sum_\ell \pi^Y_\ell\, p_\ell(x)$. Eliminating $p_k(x)$ via $\bar h^\star_k(x) \propto c_k\, p_k(x)$ yields the coordinate-wise reweighting
\begin{equation}
h^\star_k(x) \;=\; \frac{(\pi^Y_k/c_k)\, \bar h^\star_k(x)}{\sum_\ell (\pi^Y_\ell/c_\ell)\, \bar h^\star_\ell(x)},
\end{equation}
which is invertible because $\pi^Y_k/c_k > 0$ for all $k$.

Combining the two displays, $h^\star$ is a deterministic function of $g^\star$: $h^\star(x)=\psi(g^\star(x))$ for a fixed measurable map $\psi$ (the linear map $V^+$ followed by the coordinate-wise reweighting). The tower property then makes the sufficiency claim precise: for every $k$,
\begin{equation*}
\E\bigl[\mathbf{1}\{Y=k\}\mid g^\star(X)\bigr]
=\E\bigl[\underbrace{\E[\mathbf{1}\{Y=k\}\mid X]}_{=\,h^\star_k(X)=\psi_k(g^\star(X))}\ \big|\ g^\star(X)\bigr]
=\psi_k(g^\star(X))=\Pp(Y=k\mid X)\quad\text{a.s.},
\end{equation*}
so $\Pp(Y=k\mid X)=\Pp(Y=k\mid g^\star(X))$ almost surely, i.e.\ $Y\perp X\mid g^\star(X)$. Hence any Bayes-optimal classifier of $Y$ from $X$ can be expressed as a function of $g^\star(X)$ without loss; in this Bayes/predictive sense, $g^\star$ is a sufficient statistic for $Y$ given $X$.
\end{proof}

\begin{remark}[On the effective prior]
$\bar\pi$ is not an additional assumption or a free parameter: the factorization $\mu_X=\sum_k \bar\pi_k p_k$ above says precisely that $\bar\pi_k=\Pp(Y=k)$ in the pooled training population, so $\bar h^\star$ is the honest Bayes posterior of the data one actually trains on, and it is available without any prior input. External information enters only through the reweighting step, and only when predictions must be reported under a class prior $\pi^Y$ that differs from $\bar\pi$ (e.g., the class frequencies of a deployment population); such a $\pi^Y$ is not identifiable from the training data and must be supplied by the user.
\end{remark}

\subsection{Proof of Theorem~\ref{thm:identifiability} (Vertex Identifiability)}
\label{app:proofs:identifiability}

\begin{proof}
\emph{(i) Anchor points map to vertices.} Fix $k$ and let $x \in \X_k$. By \asmp{ass:irreducibility}, $p_j(x) = 0$ for $j \ne k$, and $\X_k$ has positive mass under $p_k$, so the formula in Theorem~\ref{thm:simplex} reduces to
\begin{equation}
\alpha_j(x) \;=\; \frac{c_j \cdot 0}{c_k\, p_k(x)} \;=\; 0 \quad (j \ne k), \qquad \alpha_k(x) \;=\; \frac{c_k\, p_k(x)}{c_k\, p_k(x)} \;=\; 1.
\end{equation}
Hence $g^\star(x) = v_k$ for $\mu_X$-a.e.\ $x \in \X_k$.

\emph{(ii) Vertices are extreme points of $\conv \supp(g^\star_\#\mu_X)$.} Let $\mathcal{G} := \supp(g^\star_\#\mu_X) \subseteq \Delta^{M-1}$ be the set of all mixture classifier outputs. We show $\mathcal{G} \subseteq \conv\{v_1,\dots,v_K\}$ and $v_k \in \mathcal{G}$ for each $k$, then conclude using affine independence.

\textbf{Containment.} By Theorem~\ref{thm:simplex}, $g^\star(x) \in \conv\{v_1,\dots,v_K\}$ for $\mu_X$-a.e.\ $x$, and $\conv\{v_1,\dots,v_K\}$ is closed; the support of the pushforward is contained in this closed set, so $\mathcal{G} \subseteq \conv\{v_1,\dots,v_K\}$.

\textbf{Each vertex is hit.} The marginal mass on $\X_k$ is
\begin{equation}
\mu_X(\X_k) \;=\; \frac{1}{M}\sum_m q_m(\X_k) \;=\; \frac{1}{M}\sum_m \pi_{mk}\, \Pp(X\in\X_k \mid Y=k) \;=\; \frac{c_k}{M}\, \Pp(X\in\X_k \mid Y=k) \;>\; 0,
\end{equation}
where the second equality uses \asmp{ass:irreducibility} (only class $k$ contributes on $\X_k$) and the inequality uses that $\X_k$ has positive class-conditional mass. Combined with~(i), $v_k \in \mathcal{G}$.

\textbf{Extreme points.} By Theorem~\ref{thm:simplex} the $v_k$ are affinely independent, so each $v_k$ is an extreme point of $\conv\{v_1,\dots,v_K\}$. Since $\{v_1,\dots,v_K\} \subseteq \mathcal{G} \subseteq \conv\{v_1,\dots,v_K\}$, the convex hulls of $\mathcal{G}$ and of $\{v_1,\dots,v_K\}$ coincide, and the extreme points of this common hull are exactly $\{v_1,\dots,v_K\}$.

\emph{Identifiability of $\Pi$.} Extreme points of a finite-dimensional convex set are determined by the set, hence by the population law of $g^\star(X)$. Recovering $\{v_1,\dots,v_K\}$ as an unordered set determines $\{\Pi_{:,k} = c_k v_k\}$ once the abundances are known, and these are pinned down exactly: row-stochasticity of $\Pi$ reads, row by row, $\sum_k (v_k)_m\, c_k = \sum_k \pi_{mk} = 1$, i.e.\ $c$ solves the linear system $Vc=\mathbf{1}_M$, whose solution is unique because $\rank(V)=K$ (so $c=V^+\mathbf{1}_M$). The familiar constraint $\sum_k c_k = M$ is then automatic rather than an input: multiplying $Vc=\mathbf{1}_M$ by $\mathbf{1}_M^\top$ and using that the columns of $V$ sum to one gives $\sum_k c_k = M$. The labelling of which vertex corresponds to which class index is not identifiable.

\end{proof}

\section{Latent Class Cardinality (K) Discovery}
\label{app:true_K_Discovery}
\begin{figure}
     \centering
     \includegraphics[width=1\linewidth]{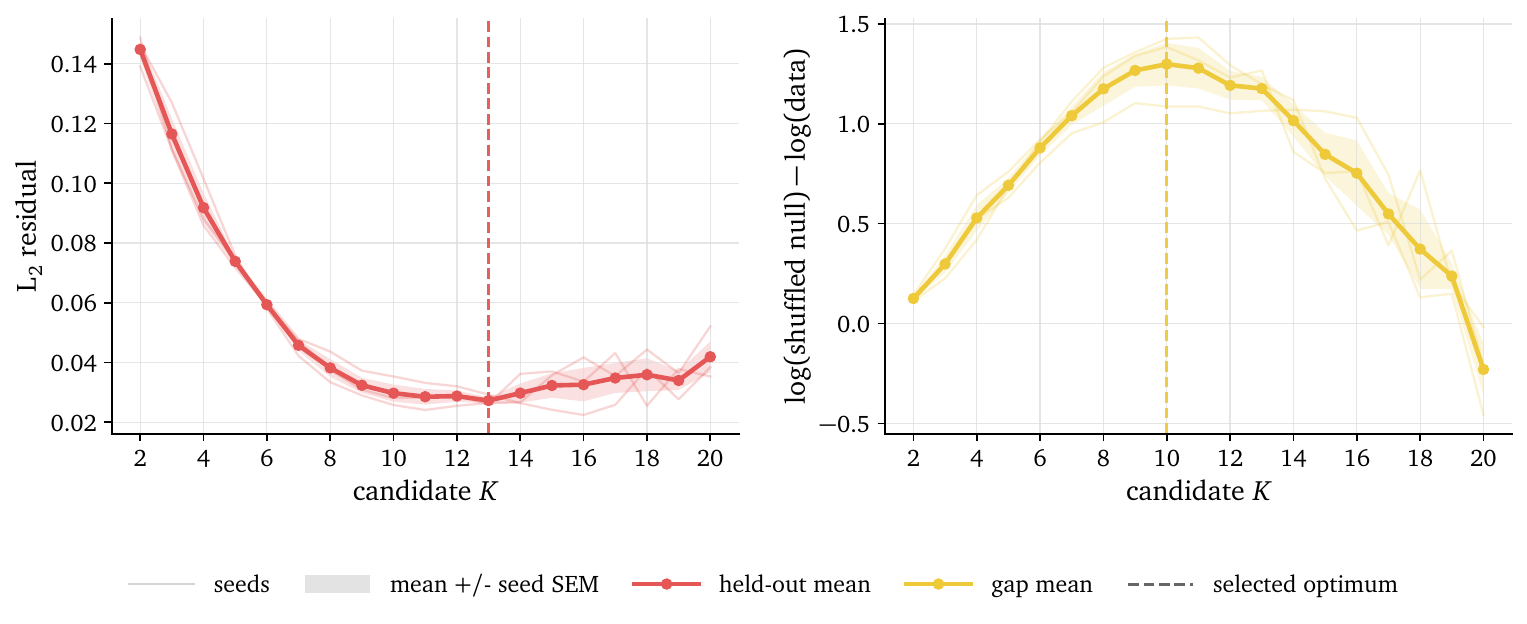}
     \caption{Latent Class Cardinality Fitting on the Galaxy10 DECaLS dataset for $K_{\text{true}}=10$ classes and $M=20$ mixtures. The left plot shows the fitted latent class cardinality using the reconstruction error on a held-out validation sample, which overestimates $K_{\text{true}}$. The right plot shows the gap statistic, which recovers $K_{\text{true}}$.}
     \label{fig:vertex_discovery}
 \end{figure}

Once the multiclass CWoLa classifier has been trained and the posteriors for all data points are computed, one does not need to know the true number of latent classes, but can discover them through the topology of the data points. 
In practice, to determine the correct cardinality without prior knowledge, we fit a $(K-1)$-simplex for each candidate $K$ within the range $[2,M]$. To counter the trivial complexity bias that naturally drives reconstruction error down as $K$ increases, we combine complementary selection signals: (1) \emph{held-out reconstruction error} scored on a validation split after fitting on the training split, which ceases to improve once $K > K_{\text{true}}$ as extra vertices begin fitting training noise, and (2) a \emph{gap statistic} computed against $B$ extra simplices fit on column-wise-shuffled posteriors (destroying joint structure while preserving marginals). \\ 
Empirically, this joint strategy successfully discovers the true latent class cardinality. We demonstrate this robust recovery on the Galaxy10 dataset with $M=20$ mixtures across three separate random seeds (see Figure~\ref{fig:vertex_discovery} for the individual seed trajectories). we also evaluated it on the MNIST dataset across a wide sweep of configurations, examining multiple target classes ($K \in \{3,5,7,10\}$) under balanced ($M=K$), moderately overcomplete ($M=K+2$), and highly overcomplete ($M=2K$) mixture conditions. 
Qualitatively, the results on MNIST confirm the reliability of the approach. In the balanced setting ($M=K$), both the held-out validation error and the gap statistic are precise, identifying the exact true cardinality across almost all seeds. In the more complex overcomplete regimes ($M > K$), the held-out validation error occasionally exhibits a mild tendency to overestimate $K$ as the model fits residual noise, while, the column-shuffled gap statistic slightly underestimated $K_{\text{true}}$. 
\section{Implementation details}
\label{app:experimental-details}
 
\paragraph{Mixture instantiation.}
After fixing $\Pi$, each mixture $\D_m$ is generated by drawing
$y\sim\pi_m$ and then sampling an example uniformly from the class-$y$
training pool. Sampling is independent across mixtures, so the same
underlying image may appear in multiple mixtures. Only $m$ is observed by
prior-free methods; ground-truth labels are used solely for permutation
alignment and reporting.
 
\paragraph{Mixture construction.}
Regular mixture construction draws each row from a symmetric Dirichlet. If $M \ge K$, each Dirichlet row is blended with a fixed cyclic identity-like template 
\begin{equation}
    \Pi_{\mathrm{final}} = (1-\lambda)\Pi_{\mathrm{Dir}} + \lambda C,
\end{equation}
and then renormalized. Purity sweeps use \textit{cyclic purity} mode, in which the dominant class receives weight $\rho$ and the remaining $1-\rho$ mass is distributed equally across the other $K-1$ classes. We use $\lambda=0.35$ for MNIST/Fashion-MNIST/CIFAR-10 and $\lambda=0.25$ for Galaxy10. We confirm in Fig.~\ref{fig:purity} and Fig.~\ref{fig:purity_cifar} that performance remains stable when the abundance of the dominant class per mixture satisfies $\max_k \pi_{mk}>0.4$.
 
\paragraph{Alignment and metrics.}
Latent class identities are recoverable only up to permutation. We compute
the Hungarian assignment between predicted and true classes on a held-out
split, and report \emph{aligned accuracy} after this assignment. ECE is
computed on the same aligned predictions. The same protocol is used for
all methods whose outputs are permutation-invariant.
 
\paragraph{Baselines.}
\textbf{Supervised oracle}: trained directly on $y$.
\textbf{Oracle simplex}: vertex hunting on $g_\theta$ with $\Pi$
known. \textbf{Wei-CCM/RCM}~\citep{wei2024consistent} and the
\citep{Tang2022MulticlassCF} demixer are known-prior multiclass methods.
\textbf{OvR}~\citep{Rifkin2004InDO} is the only prior-free baseline besides
ours; it uses the same backbone and training budget. Our prior-free
configurations (Bottleneck and Simplex) observe only $m$.
 
\paragraph{Hyperparameter control.}
Within each dataset, all compared methods share the same backbone,
optimiser (AdamW), cosine-with-warm-up schedule, validation criterion
(mixture-classification accuracy on a held-out 10\% split for prior-free
methods; class accuracy on the same split for the oracle), and stopping
rule. Only the demixing component differs.
 
\paragraph{Prior-free simplex fitting configurations.}
Table~\ref{tab:prior-free-configurations} summarises the simplex fitting
mechanisms. Post-hoc variants connect to classical simplex/vertex
methods~\citep{cutler1994archetypal,Arora2012LearningTM,Chan2009ASF};
the regularised and bottleneck variants instantiate the posterior-simplex
factorisation of \S\ref{sec:simplex-geometry}. The
``Simplex'' configuration in Tab.~\ref{tab:method_comparison} and
Fig.~\ref{fig:purity} corresponds to ``Regular + constr.~$\hat\Pi$.''
 
\begin{table}[h]
\centering
\small
\setlength{\tabcolsep}{4pt}
\renewcommand{\arraystretch}{1.12}
\begin{tabularx}{\linewidth}{@{}p{0.22\linewidth}p{0.34\linewidth}X@{}}
\toprule
Configuration & Mechanism & Closest related family \\
\midrule
Regularizer
  & Training-time regularisation
  & This work \\
Regular + constr.~$\hat\Pi$
  & Post-hoc $\hat\Pi$-constrained simplex fit
  & Simplex unmixing~\citep{Chan2009ASF} \\
Regular + arch.~corner
  & Post-hoc anchor-pool vertex recovery
  & Anchor/vertex methods~\citep{Arora2012LearningTM} \\
Regular + arch.~spread
  & Post-hoc archetypal simplex fit
  & Archetypal analysis~\citep{cutler1994archetypal} \\
Bottleneck
  & Architectural factorisation
  & This work \\
\bottomrule
\end{tabularx}
\caption{Prior-free recovery configurations. Post-hoc variants fit the
simplex after training, whereas regularised and bottleneck variants impose
the geometry during training. Arch.~corner and arch.~spread use the same
ALS refinement but differ in initialisation: low-entropy near-pure posterior
points versus radial extremes of the posterior cloud.}
\label{tab:prior-free-configurations}
\end{table}

\section{Results details }
\begin{figure}[h]
    \centering
    \includegraphics[width=\textwidth]{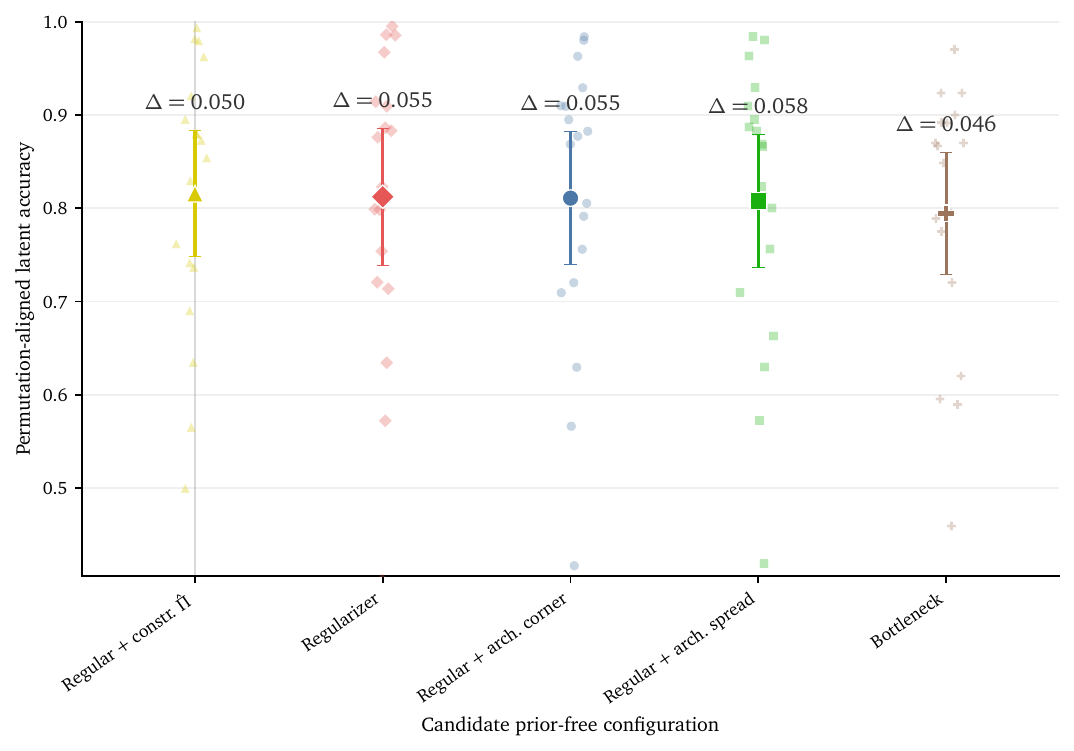}
    \caption{
    Aggregate robustness of candidate prior-free recovery configurations.
    Each row corresponds to one recovery strategy, averaged over MNIST, Fashion-MNIST, and CIFAR-10 across the evaluated $(K,M)$ settings.
    Small points show individual dataset--configuration means, while large markers show the overall mean with standard deviation.
    Text labels report the mean accuracy gap to the known-prior demixing oracle,
    $\Delta =
    \mathrm{Acc}_{\mathrm{known\text{-}prior}} -
    \mathrm{Acc}_{\mathrm{method}}$,
    where smaller is better.
    The overlapping intervals indicate that performance is robust to the precise recovery choice.
    }
    \label{fig:fitter-robustness}
\end{figure}

Figure~\ref{fig:fitter-robustness} compares simplex fitters across MNIST, Fashion-MNIST, and CIFAR-10.
All configurations remain close to the known-prior demixing oracle, with mean gaps between $0.046$ and $0.058$, and their confidence intervals largely overlap.
No single fitter uniformly dominates, indicating that the posterior-simplex signal is stable across several recovery mechanisms.

\begin{figure}[h]
    \centering
    \includegraphics[width=\linewidth]{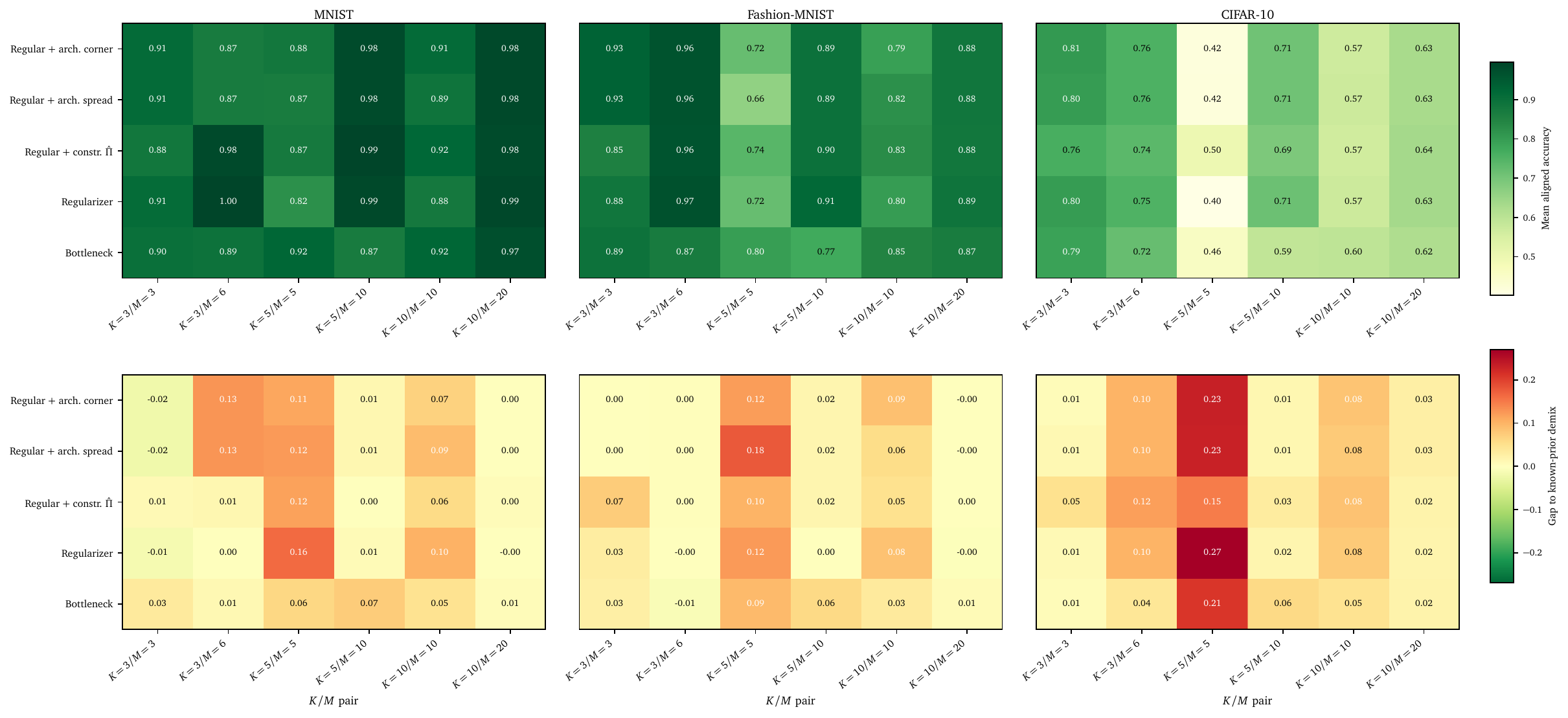}
    \caption{
    Per-dataset decomposition of the prior-free recovery comparison.
    \emph{Top:} mean permutation-aligned latent accuracy for each
    recovery configuration on MNIST, Fashion-MNIST, and CIFAR-10.
    \emph{Bottom:} accuracy gap to the known-prior demixing reference,
    $\Delta=\mathrm{Acc}_{\text{known-prior}}-\mathrm{Acc}_{\text{method}}$
    (positive = the method is worse than the known-prior reference).
    Columns vary $(K,M)$. Square settings $K=M=5$ are the most fragile,
    most clearly on CIFAR-10; overcomplete settings $M>K$ shrink the
    oracle gap, consistent with the redundancy argument in Sec.~\ref{sec:simplex-geometry}.}
    \label{fig:heatmap-fitters}
\end{figure}

In this work we measure the distance to the oracle simplex using 
\begin{equation*}
    d_{\text{oracle}} = \sqrt{\frac{1}{K}\min_{\sigma \in S_K} \sum_{k=1}^{K} \|\hat{\theta}_k - \theta_{\sigma(k)}\|^2}.
\end{equation*} Figure~\ref{fig:heatmap-fitters} shows that variation is driven more by the mixture regime than by the fitter itself.
The square settings $K=M$ are often the most fragile, especially near $K=M=5$, whereas overcomplete settings with $M>K$ provide redundant geometric constraints and are typically easier.
This robustness makes post-hoc fitters practically useful: after training one mixture classifier, several simplex recovery rules can be tested cheaply on the same posterior cloud, without changing the architecture or retraining.

\label{app:results-details}

\begin{figure}[h]
    \centering
    \includegraphics[width=1\textwidth]{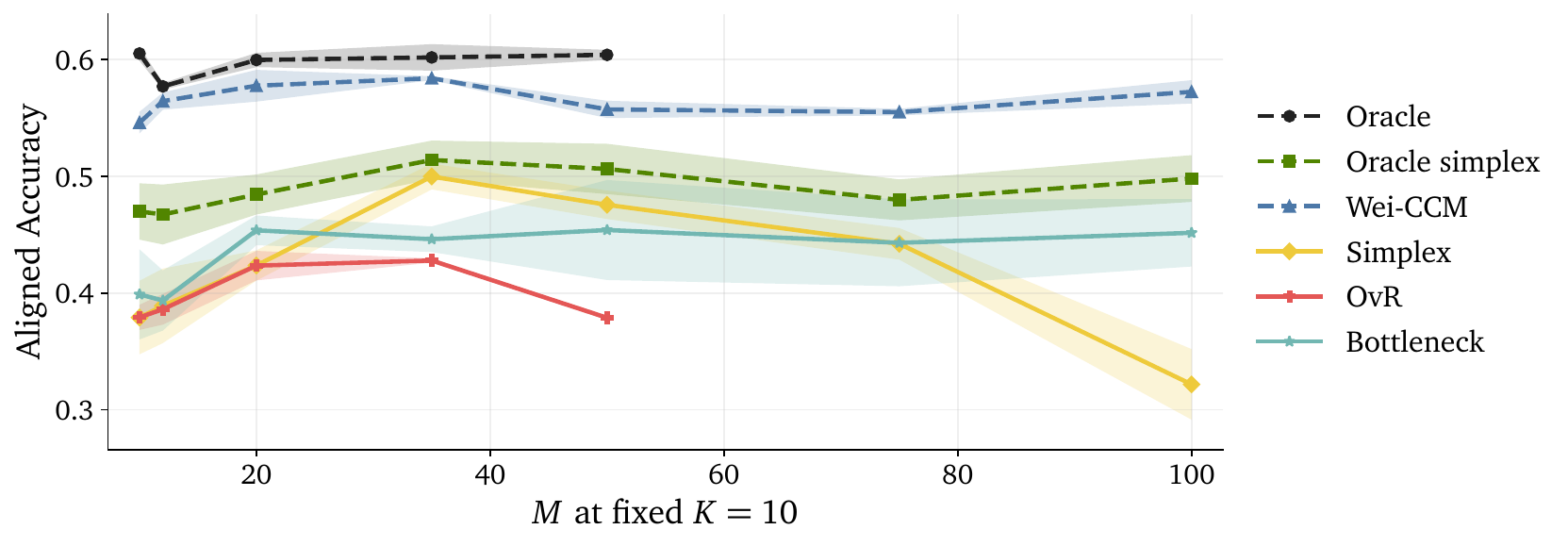}
    \caption{
    Scaling with the number of mixtures on Galaxy10 DECaLS.
    We fix the number of latent classes to $K=10$ and increase the number of mixtures $M$ while keeping the image data, backbone, and augmentation profile fixed.
    Larger values of $M$ provide additional mixture diversity and improve identifiability of the latent simplex structure.
    Performance generally improves as the number of mixtures increases, particularly for simplex-based approaches.
    Our simplex and bottleneck variants consistently outperform the no-prior OvR baseline and approach the known-prior simplex and Wei-CCM references at larger $M$.
    For computational reasons, the OvR and fully supervised oracle baselines are only evaluated up to $M=50$.
    }
    \label{fig:gal_m_sweep}
\end{figure}

Figure~\ref{fig:gal_m_sweep} extends the fixed-$K$ scaling study to Galaxy10. 
As on CIFAR-10, increasing the number of mixtures generally improves the quality of prior-free recovery, consistent with the view that additional mixtures provide extra geometric constraints on the latent simplex. 
The gain is clearest for the simplex-based methods, which remain above the no-prior OvR baseline and move closer to the known-prior references at larger $M$.

\begin{figure}[h]
    \centering
    \includegraphics[width=1\textwidth]{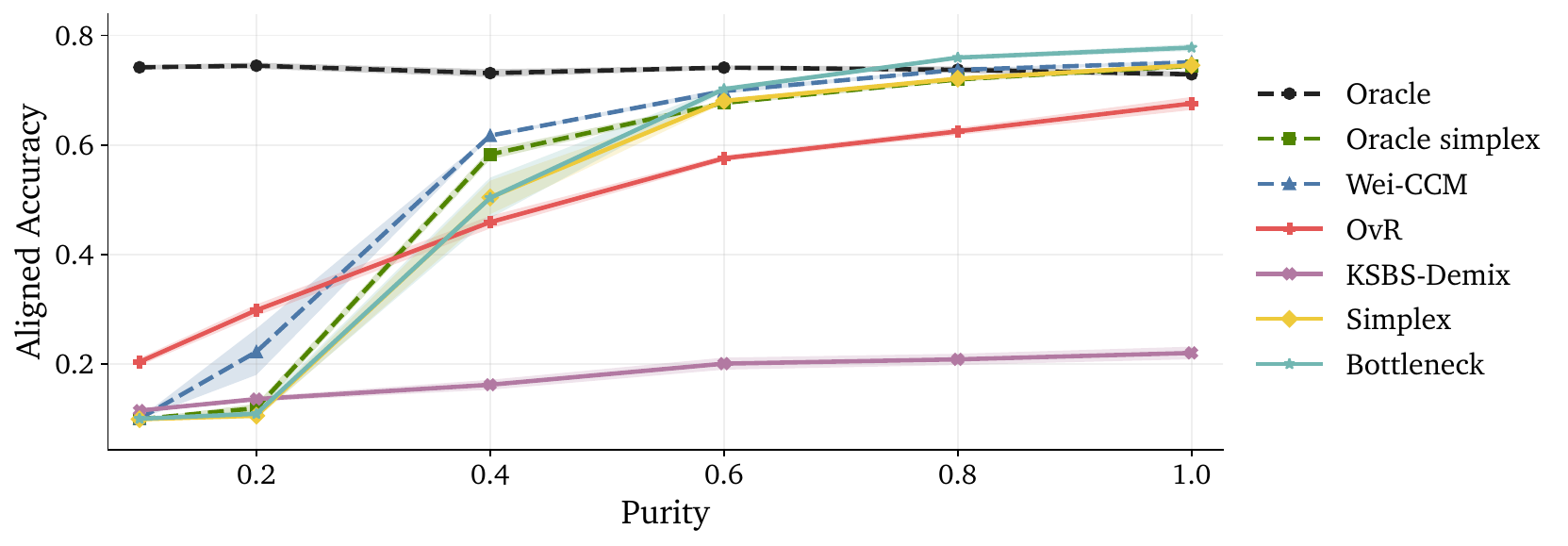}
    \caption{
    Robustness to mixture purity on CIFAR-10.
    We construct cyclic-purity mixtures with $K=M=10$ while keeping the image data and training protocol fixed.
    Low purity corresponds to weak class enrichment and approaches the unidentifiable regime, whereas high purity produces mixtures with clearer anchor structure.
    Accuracy improves with purity across all mixture-based methods.
    Our simplex-based approach consistently outperforms the no-prior OvR baseline at moderate and high purity, while approaching the known-prior simplex and Wei-CCM reference methods in the high-purity regime.
    }
    \label{fig:purity_cifar}
\end{figure}

Figure~\ref{fig:purity_cifar} isolates the effect of mixture informativeness on CIFAR-10. 
When the cyclic-purity parameter is small, the mixtures are nearly indistinguishable and the weakly supervised methods approach the expected non-identifiable regime. 
As purity increases, the simplex and bottleneck variants improve rapidly and approach the known-prior baselines, indicating that the limiting factor is mixture separability rather than the simplex fitting step itself.

Table~\ref{tab:method_comparison_notune} reports the corresponding Galaxy10 comparison with a frozen backbone. 
In this lower-capacity setting, the bottleneck gives the best prior-free accuracy, while the post-hoc simplex fit gives the best prior-free calibration. 
This contrasts with the finetuned setting in Table~\ref{tab:method_comparison}, where post-hoc simplex recovery gives the strongest prior-free tradeoff, suggesting that the preferred recovery mechanism can depend on representation quality.
\begin{table*}[t]
\centering
\begin{tblr}{
  width = \textwidth,
  colspec = {X[l,m]X[c]X[c]X[c]},
  colsep = 18pt,
  stretch = 1.0,
  column{1} = {leftsep=0pt},
  column{4} = {rightsep=0pt},
}
\toprule
Technique & Method & Accuracy & ECE \\
\midrule

Supervised
& Oracle
& $0.600 \pm 0.015$
& $0.045 \pm 0.010$ \\

\midrule[dotted]

Prior-aware
& Wei CCM
& $0.578 \pm 0.033$
& $0.044 \pm 0.017$ \\

& Wei RCM
& $0.455 \pm 0.030$
& $0.270 \pm 0.032$ \\

& Oracle Simplex
& $0.484 \pm 0.042$
& $0.082 \pm 0.037$ \\

\midrule[dotted]

Prior-free
& OvR
& $0.423 \pm 0.031$
& $0.222 \pm 0.033$ \\

& Bottleneck
& $0.454 \pm 0.024$
& $0.154 \pm 0.051$ \\

& Simplex
& $0.439 \pm 0.026$
& $0.111 \pm 0.059$ \\

\bottomrule
\end{tblr}

\caption{
Galaxy10 comparison on the non fine-tune model.
Results are reported as mean $\pm$ standard deviation over three random seeds.
Accuracy is permutation-aligned latent accuracy; ECE is expected calibration error, where lower is better.
Oracle is fully supervised.
Wei CCM and Wei RCM assume access to class-prior information~\citep{wei2024consistent}, and Oracle Simplex uses the true mixture proportions.
OvR~\citep{Rifkin2004InDO}, Bottleneck, and Simplex are prior-free.
Bottleneck achieves the strongest prior-free accuracy, while Simplex provides the best calibration among prior-free methods.
}
\label{tab:method_comparison_notune}

\end{table*}
\newpage

\end{document}